\documentclass{article} 
\usepackage[utf8]{inputenc}
\PassOptionsToPackage{table,xcdraw,usenames,dvipsnames}{xcolor}
\usepackage[T1]{fontenc}
\DeclareUnicodeCharacter{266B}{\ding{72}}
\usepackage{iclr2027_conference,times}
\usepackage[hidelinks]{hyperref}
\usepackage{hyperref}
\usepackage{url}
\usepackage{graphicx}
\usepackage{adjustbox}
\usepackage{booktabs}
\usepackage{pifont}
\usepackage{tabularx}
\usepackage{multirow}
\usepackage{caption}
\usepackage{siunitx}
\usepackage{array}
\usepackage{pdfpages}
\usepackage{longtable}
\usepackage{amsmath}
\usepackage{cleveref}
\usepackage{appendix}
\usepackage{changes}
\usepackage{amsthm}
\usepackage{xcolor}
\usepackage{mdframed}
\usepackage{thmtools}
\usepackage{thm-restate}
\usepackage[most]{tcolorbox}
\usepackage{amssymb}
\usepackage{placeins}
\usepackage{float}
\usepackage{xspace}

\usepackage{xcolor}

\definecolor{ForestGreen}{RGB}{34,139,34}
\definecolor{myyellow}{RGB}{181, 181, 27}

\renewcommand{\thefootnote}{\fnsymbol{footnote}}

\newcommand{\brandicon}[2][0.9em]{%
  \raisebox{-0.15\height}{%
    \includegraphics[height=#1]{#2}%
  }%
}
\newcommand{\qwen}{%
  \hspace{-0.2em}%
  \raisebox{-0.3\height}{%
    \includegraphics[height=1.3em]{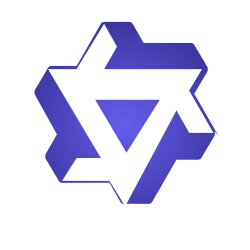}%
  }%
}

\newcommand{\microsoft}{%
  \raisebox{-0.15\height}{%
    \includegraphics[height=0.9em]{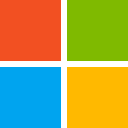}%
  }\hspace{0.23em}%
}

\newcommand{\google}{%
  \hspace{-0.15em}%
  \raisebox{-0.2\height}{%
    \includegraphics[height=1.3em]{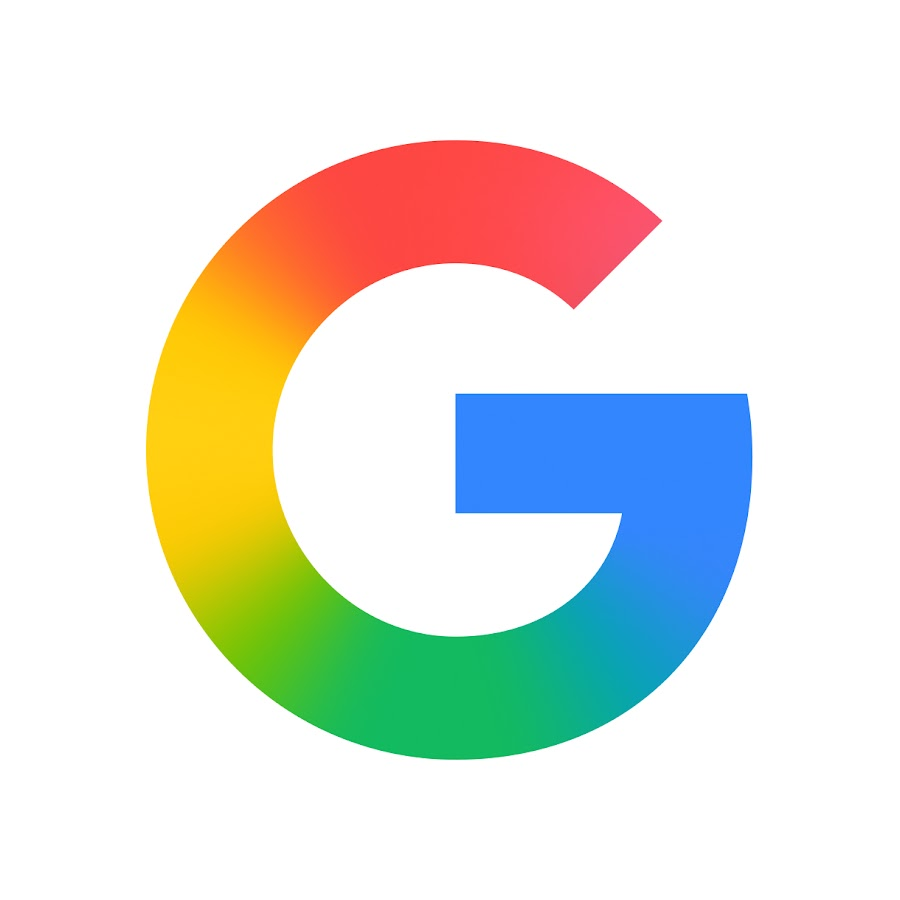}%
  }%
}

\newcommand{\meta}{%
  \hspace{-0.15em}%
  \raisebox{-0.2\height}{%
    \includegraphics[height=1.1em]{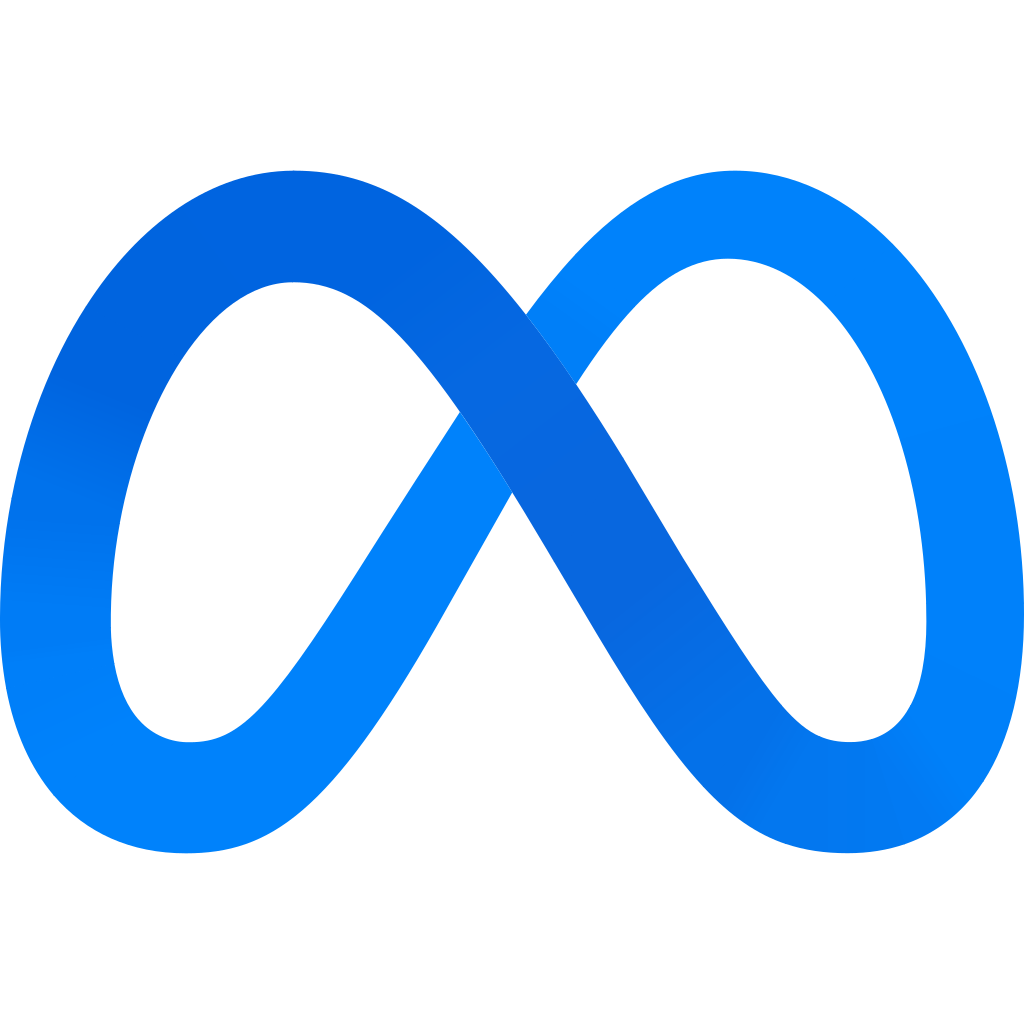}%
  }\hspace{0.2em}%
}

\newcommand{\deepseek}{%
  \hspace{-0.15em}%
  \raisebox{-0.2\height}{%
    \includegraphics[height=1.1em]{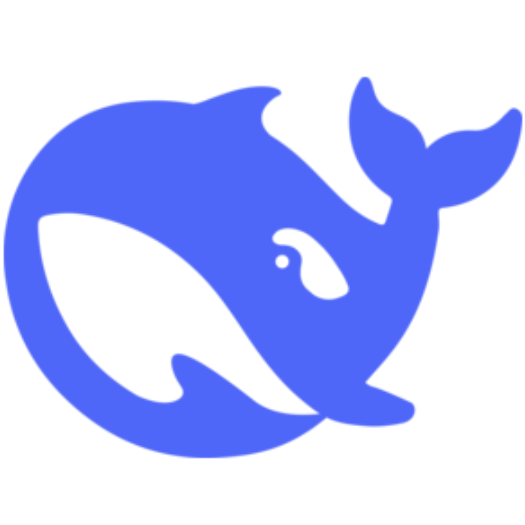}%
  }\hspace{0.15em}%
}

\newcommand{\ministra}{%
  \hspace{-0.15em}%
  \raisebox{-0.2\height}{%
    \includegraphics[height=1.15em]{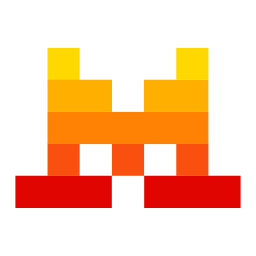}%
  }\hspace{0.15em}%
}

\usepackage{makecell}
\usepackage{enumerate}
\usepackage{enumitem}

\definecolor{mygrey}{gray}{0.4}

\usepackage[ruled,algo2e,vlined]{algorithm2e}
\SetKwInOut{Input}{Input}\SetKwInOut{Output}{Output}
\SetKwComment{Comment}{$\triangleright$\ }{}

\usepackage{listings}

\definecolor{shadecolor}{gray}{0.92}
\declaretheoremstyle[
  headfont=\normalfont\bfseries,
  notefont=\mdseries,
  notebraces={(}{)},
  bodyfont=\normalfont,
  postheadspace=0.5em,
  spaceabove=6pt,
  spacebelow=6pt,
  mdframed={
    skipabove=8pt,
    skipbelow=8pt,
    hidealllines=true,
    backgroundcolor=shadecolor,
    innerleftmargin=4pt,
    innerrightmargin=4pt,
    innertopmargin=4pt,
    innerbottommargin=4pt
  }
]{shaded}

\usepackage[most]{tcolorbox}

\newcolumntype{L}[1]{>{\raggedright\arraybackslash}p{#1}}

\definecolor{RowHL}{RGB}{215,244,250} 

\usepackage{amsmath,amsfonts,bm}

\def\eqref#1{equation~\ref{#1}}

\def\1{\bm{1}}

\DeclareMathAlphabet{\mathsfit}{\encodingdefault}{\sfdefault}{m}{sl}
\SetMathAlphabet{\mathsfit}{bold}{\encodingdefault}{\sfdefault}{bx}{n}

\title{BaRe-Mem: \textbf{Ba}yesian \textbf{Re}liability \textbf{Mem}ory for Robust and Adaptive Agent Consultation}

\iclrfinalcopy

\author{
Peilin Feng\textsuperscript{1}
\quad
Zhengyang Huang\textsuperscript{2}
\quad
Soujanya Poria\textsuperscript{1,\textdagger}
\\[4pt]
\textsuperscript{1}DeCLaRe Lab, Nanyang Technological University
\quad
\textsuperscript{2}Peking University
}

\begin{document}

\maketitle

\vspace{-3.0em}

\begin{center}
\begin{tabular}{ll}
\raisebox{-0.15em}{\includegraphics[height=1.05em]{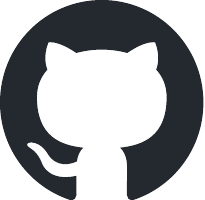}}
\textbf{Github:}
&
\url{https://github.com/declare-lab/BaRe-Mem}
\\

\raisebox{-0.15em}{\includegraphics[height=1.05em]{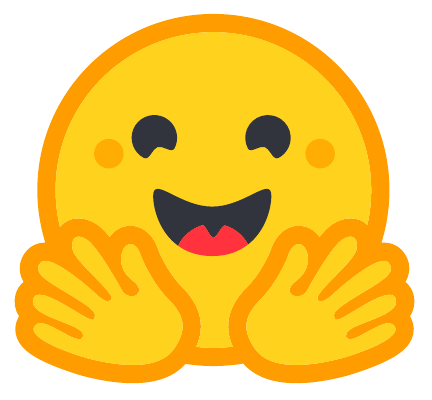}}
\textbf{Dataset:}
&
\url{https://huggingface.co/datasets/Sssunset/BaRe-Mem-Data}
\\

\end{tabular}
\end{center}

\begingroup
\renewcommand{\thefootnote}{\textdagger}
\footnotetext{Corresponding author. E-mail: \texttt{soujanya.poria@ntu.edu.sg}}
\endgroup

\begin{abstract}
In multi-agent systems, reliable consultation is challenging because advisor capabilities vary across tasks, and misleading information can make consultation worse than autonomous reasoning. We introduce \textbf{BaRe-Mem}, an online Bayesian reliability memory for multi-agent consultation. It estimates advisor reliability based on the central model's internal belief representations and updates these estimates from historical interactions. These estimates modulate the influence of advisor responses and guide the choice between consultation and autonomous reasoning. Across nine benchmarks and six central models, BaRe-Mem is more robust to misleading advisor information than debate and majority voting. On the more challenging tasks, it remains above autonomous reasoning across all tested misleading levels. Moreover, we extend the BaRe-Mem mechanism to worker allocation in agent teams. On the MuSiQue benchmark, BaRe-Mem improves task completion over routing by historical success counts and identifies capable workers earlier.
\end{abstract}

\vspace{-1.5em}
\section{Introduction}
\vspace{-0.5em}
As no single large language model (LLM) can be expected to possess all the capabilities required for complex tasks~\citep{shen2023hugginggpt, jiang2023llm,wang2025mixture}, AI systems are increasingly being organized as networks of interacting agents~\citep{wu2023autogen, guo2024large}. These agents may consult other models~\citep{shen2023hugginggpt}, specialized services~\citep{song2023restgpt}, tools~\citep{qin2024toolllm}, or humans~\citep{wu2023autogen}, drawing on capabilities that lie outside their own parameters. This makes reliable consultation a fundamental problem: \textit{an agent must determine not only what other sources provide but also how much their information should influence its own decision.}

Reliable consultation is necessary because advisor information can help as well as hurt~\citep{du2023improving,cui2026free}. Recent work shows that exposure to misleading consultation can cause models to abandon initially correct answers, which reduces the performance of the agent seeking advice~\citep{song2025llmscanthandlepeer}. This risk is compounded by the heterogeneous and task-dependent capabilities of advisors: a model that is reliable on one domain may fail on another~\citep{smit2023should,kim2026capable}, while agreement among multiple advisors does not guarantee correctness when they reinforce the same plausible but erroneous solution~\citep{weng2025we,zhu2025conformity}. Effective consultation therefore requires more than aggregating current responses: \textit{the central model must estimate which evidence in external information is reliable for the current question.}

Repeated interaction history naturally provides such information. Existing methods preserve past experience either as textual memory, such as retrieved trajectories~\citep{zhao2024expel}, distilled summaries~\citep{NEURIPS2025_136a45cd}, and working skills~\citep{yu2026recursive}, or as persistent continuous states that are updated across interactions~\citep{behrouz2026titans,zhang2026memgen,feng2026sigma,bayat2026proteus}. However, only retaining history is not enough. A reliable historical memory should turn past interactions into contextual estimates that regulate the influence of external information on the current decision~\citep{teacy2006travos,zhou2026epistemic}. In order to remain effective over long horizon interactions, these estimates should scale with an expanding task stream, remain robust to misleading consultation, and adapt to changes in advisor reliability. Moreover, since collaboration can yield diminishing or even negative gains as single-agent capability becomes high enough~\citep{kim2026capable,verma2026selene}, reliability should determine not only how external evidence influences the central model, but whether it should be utilized at all~\citep{eo2025debate,verma2026selene,zhang2025stop}.  A robust reliability memory should therefore assess not only the advisors' information reliability but also the central model ability itself, enabling it to determine whether external evidence is likely to improve upon its autonomous reasoning on the current question.

To address these challenges, we introduce \textbf{BaRe-Mem}, an online Bayesian reliability memory for robust and adaptive multi-agent consultation. It models advisor reliability conditioned on the central model's internal belief representations of the current question and candidate response. The memory maintains these estimates for both the central model and its advisors, updating them online from verified correctness outcomes. Advisor reliability estimates steer attention to peer responses. Additionally, it compares estimated consultation and autonomous ability to decide whether to consult. In the experiments, we show that \textbf{BaRe-Mem} remains robust as external information becomes increasingly misleading by adaptively shifting between consultation and autonomous reasoning. Its predicted consultation advantage is consistent with the real gain observed after verification, and useful reliability estimates emerge from sparse verified feedback. BaRe-Mem also improves worker selection in agent teams through verification feedback, extending its deployment scenarios beyond response level consultation.

\vspace{-1em}
\section{Related Work}
\vspace{-0.5em}
\paragraph{Memory Construction}
Existing approaches construct memory through either explicit textual records or latent continuous memory states. Text-based systems store historical information externally or distill feedback and experience into reusable reflections, skills, and trajectories~\citep{shinn2023reflexion,packer2023memgpt,zhong2024memorybank,zhao2024expel,chhikara2025mem0,NEURIPS2025_136a45cd}. Despite their flexibility, textual memories are constrained by compression fidelity and retrieval noise~\citep{laban2026llms}. In parallel, continuous memory approaches encode past experience into persistent latent states, neural memory, generated memory tokens, or structured competence states that can be updated across interactions~\citep{wu2022memorizing,wang2024memoryllm,wang2025m+,behrouz2026titans,zhang2026memgen,wei2026mlp,feng2026sigma,bayat2026proteus,cao2026memory}. However, these approaches do not explicitly model an advisor's reliability estimation conditioned on the central model's internal belief of the context.
\paragraph{Memory Steering.}
Memory can steer model behavior through three broad interfaces:
prompt conditioning, parameter adaptation, and internal-state modulation. Prompt-based approaches retrieve or summarize historical experience into textual prompts that guide subsequent
reasoning~\citep{shinn2023reflexion,zhao2024expel,zhou2026epistemic}.
Parameter-based approaches encode historical information into low-rank
adaptations, allowing memory to alter subsequent computation while keeping the pretrained backbone fixed~\citep{wang2024greater,charakorn2026doc}. Internal-state approaches instead inject memory directly into the model's computation, through recurrent matrix states~\citep{yang2024parallelizing,team2025kimi}, activation or residual modulation~\citep{lei2026delta,feng2026sigma}, or direct modification of attention logits or weights~\citep{zhang2024tell,guardieiro2025instruction,
yan2025don,deng2025cram}. We adopt attention modification because BaRe-Mem produces specific reliability estimates that can directly modulate the influence of each advisor's response.
\vspace{-1em}
\section{BaRe-Mem: Reliability-Guided Consultation}
\vspace{-0.5em}
BaRe-Mem estimates contextual reliability for both the central model and its advisors from verified interaction history. These estimates serve two roles: modulating the influence of advisor responses and determining whether consultation is preferable to autonomous reasoning.

\subsection{Organising Historical Reliability}
\label{Section 3.1}

\paragraph{Belief Representation.}
For question $q_t$, we consider $K+1$ candidates: $K$ advisor responses available to the central model $M$ for consultation and its autonomous answer $a_{t,0}$, which is used only for autonomous ability estimation. The frozen central model $M$ encodes each candidate $k$ in the context of $q_t$, yielding a hidden representation $h_{t,k}$. From these representations, we derive a question belief $\psi_q(t)$ shared across candidates and an answer content belief $\psi_c(t,k)$ specific to candidate $k$. We represent candidate $k$ as
\begin{equation}
x_{t,k} = \big[e_k \otimes \psi_q(t)\,;\,\psi_c(t,k)\,;\,1\big],
\end{equation}
where $e_k$ denotes the one-hot identity of its source. The reliability score is then predicted as
\begin{equation}
w^\top x_{t,k}
=
\underbrace{w_k^\top \psi_q(t)}_{\text{source reliability}}
+
\underbrace{w_c^\top \psi_c(t,k)}_{\text{answer content reliability}}
+
\underbrace{w_0}_{\text{bias}}
\end{equation}
The first term captures the reliability of source $k$ on questions represented similarly to $q_t$, while the second captures reliability evidence from the candidate response itself.  

\paragraph{Memory.}
BaRe-Mem models the verified correctness with Bayesian linear regression. Let
$s_{t,k}=2y_{t,k}-1$, where $y_{t,k}\in\{0,1\}$ indicates whether candidate $k$ is correct:
\begin{equation}
s_{t,k}\mid x_{t,k},w
\sim
\mathcal N\!\left(w^\top x_{t,k},1\right),
\qquad
w\sim\mathcal N\!\left(0,\lambda^{-1}I\right).
\end{equation}
Given all verified candidates observed so far, the posterior is
\begin{equation} \Lambda = \lambda I+\sum xx^\top, \qquad b = \sum s\,x, \qquad m = \Lambda^{-1}b. \end{equation}

When a new candidate is verified, this posterior can be maintained exactly through the rank-one update using kalman gain:

\begin{equation} g_{t,k} = \frac{\Lambda^{-1}x_{t,k}} {1+x_{t,k}^\top\Lambda^{-1}x_{t,k}}, \quad m \leftarrow m + g_{t,k} \big( s_{t,k}-x_{t,k}^\top m \big), \quad \Lambda^{-1} \leftarrow \Lambda^{-1} - g_{t,k} \big( \Lambda^{-1}x_{t,k} \big)^{\!\top}. \end{equation} 

The Kalman gain $g_{t,k}$ adaptively weights each verified outcome according to the current posterior uncertainty, enabling exact online updates without recomputing the posterior from scratch. The derivation is provided in Appendix~\textcolor{blue}{\textbf{\ref{apps:lemma1}}} and Appendix\textcolor{blue}{\textbf{~\ref{app:kalman_gain}}}.

\begin{figure}[!h]
\vspace{-1em}
\centering
\includegraphics[width=0.8\linewidth]{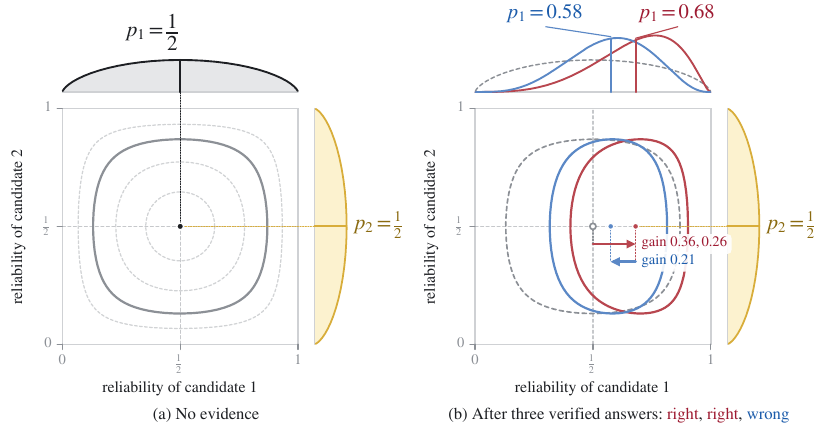}
\caption{
Illustration of the BaRe-Mem reliability memory update process.
\textbf{(a)} Without verified evidence, all advisors are assigned equal reliability.
\textbf{(b)} Updating only candidate~1 with two correct outcomes shifts its estimated reliability to the \textcolor{red}{red} posterior, while a subsequent incorrect outcome yields the \textcolor{blue}{blue} posterior. The annotated gains denote the corresponding Kalman gains.
}
\vspace{-0.8em}
\label{fig:memory_update}
\end{figure}

\paragraph{Reliability Estimate.}
\label{reliability_estimate_000}
When the central model $M$ solves question $q_t$, the reliability of candidate $k$ is estimated as its probability of being correct:
\begin{equation}
p_{t,k}
=
\Phi\!\left(
\frac{\mu_{t,k}}
{\sqrt{1+v_{t,k}}}
\right),
\qquad
\mu_{t,k}
=
x_{t,k}^\top m,
\qquad
v_{t,k}
=
x_{t,k}^\top\Lambda^{-1}x_{t,k}.
\label{esitimation}
\end{equation}
Here, $\mu_{t,k}$ is the predicted signed correctness and $v_{t,k}$ its uncertainty. The derivation can be found in Appendix~\textcolor{blue}{\textbf{\ref{app:record_estimate}}}. Figure\textcolor{blue}{\textbf{~\ref{fig:memory_update}}} illustrates this update process. Without verified evidence, the memory assigns equal reliability $p=\tfrac{1}{2}$ to all advisors. After two correct outcomes for candidate~1, its estimated reliability increases, while a subsequent incorrect outcome reduces it. Candidate~2 remains unchanged when no evidence is observed for it. The corresponding Kalman gains control the magnitude of each update.

\subsection{Reliability-Guided Attention}
\label{sec:reliability_incorporation}

We use the estimated advisor reliabilities to modulate the influence of each advisor's response on the central model. For each context token $j$ belonging to an advisor response, let $c(j)$ denote the advisor that produced that response. In every attention head\footnote{The modification is applied only to full-attention layers with softmax normalization}, we modify the attention weights as
\begin{equation}
\alpha_{qj}=
\operatorname{softmax}_j
\left(
\frac{\langle Q_q,K_j\rangle}{\sqrt d}
+
\beta_{t,c(j)}
\right),
\qquad
\beta_{t,k}=
\gamma
\log
\frac{p_{t,k}}
{\max_{k'} p_{t,k'}} .
\end{equation}

with $\beta=0$ for tokens outside advisor responses. Before softmax normalization, this rescales the unnormalized attention weight on advisor $k$ by
$\left(\frac{p_{t,k}}{\max_{k'}p_{t,k'}}\right)^\gamma \in (0,1]$.
Thus, the most reliable advisor is left unchanged, while less reliable advisors are progressively downweighted.

This attention steering introduces no trainable parameters or additional training and depends only on the relative reliability among advisors. 

\subsection{Deciding Whether to Consult}
\label{sec:whether_to_consult}
Reliability-guided attention in Section\textcolor{blue}{\textbf{~\ref{sec:reliability_incorporation}}} determines how strongly each advisor should influence the central model, but not whether consultation is preferable to autonomous reasoning. We therefore estimate both abilities on the current question and select the mode with higher estimated accuracy.

\paragraph{Consultation Ability.}
Let $T_t$ denote the highest reliability estimate among the advisor candidates, representing the memory's confidence that trustworthy evidence is available among the consulted responses. When $T\to1$, consultation succeeds with probability $\rho$, which describes how well $M$ uses trustworthy evidence. When $T\to0$, unreliable evidence may pull $M$ away from its autonomous judgment, reducing its accuracy from its autonomous ability $\kappa$ by $\delta$. We model $M$'s consultation ability by interpolating between these two regimes:
\begin{equation}
A(T)
=
\underbrace{T\,\rho}_{\text{reliable evidence regime}}
+
\underbrace{(1-T)\,(\kappa-\delta)}_{\text{unreliable evidence regime}}.
\label{eq:consultation_ability}
\end{equation}
Here, $\kappa$ is the central model's autonomous ability on the current question, independent of external evidence. Accordingly, $\kappa-\delta$ represents its consultation ability under unreliable evidence, with $\delta$ measuring the degradation relative to autonomous reasoning. Meanwhile, $\rho$ denotes the consultation ability when trustworthy evidence is available. Therefore, $A(T)$ describes the central model's expected consultation ability as a function of the memory's confidence in the external evidence.

\paragraph{Estimating Consultation and Autonomous Ability.}
\label{Estimating_ability}
For the current question $q_t$, the specific quantities $\kappa_t$ and $T_t$ are directly available from the reliability memory. The autonomous candidate provides the central model's autonomous ability $\kappa_t$, while the highest reliability among the advisor candidates gives the trust $T_t$:
\begin{equation}
\kappa_t
=
p_{t,0},
\qquad
T_t
=
\max_k p_{t,k},
\end{equation}
The consultation parameters $\rho$ and $\delta$, in contrast, are unknown and are learned from previously verified interactions. Rearranging Eq.\textcolor{blue}{\textbf{~(\ref{eq:consultation_ability})}} yields a linear form for $\theta=[\rho,\delta]^\top$:
\begin{equation}
z_t
\equiv
y_t-(1-T_t)\kappa_t,
\qquad
z_t \mid u_t,\theta
\sim
\mathcal N\!\left(u_t^\top\theta,1\right),
\qquad
u_t
=
\begin{bmatrix}
T_t\\
T_t-1
\end{bmatrix}.
\end{equation}
Here, $y_t\in\{0,1\}$ indicates whether consultation produces the correct answer on question $q_t$ and becomes available after verification. Using the same online Bayesian regression as the reliability memory\textcolor{blue}{\textbf{~\ref{apps:lemma1}}}, we estimate $\rho$ and $\delta$ from previously verified questions:
\begin{equation}
\begin{bmatrix}
\hat\rho\\
\hat\delta
\end{bmatrix}
=
P_c^{-1}q_c,
\qquad
P_c
=
I+\sum_{s<t}u_su_s^\top,
\qquad
q_c
=
\theta_0+\sum_{s<t}u_sz_s.
\end{equation}
Substituting $\kappa_t$, $T_t$, $\hat\rho$, and $\hat\delta$ into Eq.\textcolor{blue}{\textbf{~(\ref{eq:consultation_ability})}} gives the estimated consultation ability for the current question. The detailed derivation is in Appendix~\textcolor{blue}{\textbf{\ref{app:consultation_parameter_estimation}}}

\paragraph{Decision Rule.}
Given the estimated consultation and autonomous abilities, $M$ selects the mode with higher estimated accuracy:
\begin{equation}
\hat a_t
=
\begin{cases}
\text{consultation}, & \text{if } A(T_t)\ge\kappa_t,\\
\text{autonomous reasoning}, & \text{otherwise}.
\end{cases}
\label{eq:action_selection}
\end{equation}

The advantage of consultation over autonomous reasoning can be written as
\begin{equation}
A(T_t)-\kappa_t
=
\underbrace{T_t\,(\hat\rho-\kappa_t)}_{\text{gain from reliable evidence}}
-
\underbrace{(1-T_t)\,\hat\delta}_{\text{loss from unreliable evidence}}.
\label{eq:consultation_gain}
\end{equation}
Consultation is therefore preferred when the expected gain from reliable external evidence outweighs the potential degradation from unreliable evidence.

When $\hat\delta>0$ and $\hat\rho>\kappa_t$, this condition yields a decision threshold $T_t^*
= \frac{\hat\delta}{\hat\rho+\hat\delta-\kappa_t}$. $M$ consults when $T_t\ge T_t^*$. The threshold increases with autonomous ability $\kappa_t$ and the degradation $\hat\delta$, and decreases with the consultation ability under reliable evidence $\hat\rho$. Thus, a stronger autonomous model requires more trustworthy external evidence before consultation becomes preferable. The remaining parameter regimes are analyzed in Appendix\textcolor{blue}{\textbf{~\ref{app:break_even}}}.
\vspace{-0.5em}
\section{Experiments}
\subsection{Experimental Setups}
\label{experimental_setups}
We conduct our study under two complementary capability regimes. The \textbf{\textit{capability-supported}} suite includes mathematical reasoning (GSM8K~\citep{cobbe2021training}), code generation (APPS~\citep{hendrycks2021measuring}), and retrieval-based question answering (SQuAD~\citep{rajpurkar2016squad}), where most evaluated central models exhibit relatively strong competence and useful advisor information is broadly available. The \textbf{\textit{capability-challenging}} suite includes physical commonsense reasoning (PIQA~\citep{bisk2020piqa}), broad knowledge (MMLU~\citep{hendrycks2020measuring}), science question answering (OpenBookQA~\citep{mihaylov2018can} and SciQ~\citep{welbl2017crowdsourcing}), complex reasoning (BBH~\citep{suzgun2023challenging}), and language understanding (SuperGLUE~\citep{wang2019superglue}), where capabilities are substantially more heterogeneous and task-dependent across models. We utilize six advisors and evaluated six central models, their individual performance is reported in Appendix\textcolor{blue}{\textbf{~\ref{app:model_performance}}}. Within each capability regime, questions from the constituent datasets are randomly shuffled to prevent the memory from exploiting dataset order as a shortcut for advisor reliability. In the main text, we use \qwen Qwen3-14B~\citep{yang2025qwen3} and \brandicon{figures/logos/microsoft.png}\ Phi-4~\citep{abdin2024phi} as representative central models for analysis. Additional results are provided in the Appendix \textcolor{blue}{\textbf{\ref{app:additional_experiments}}}.

\subsection{Adaptive Consultation under Misleading Information}
\label{Adaptive Consultation under Misleading Information}
In heterogeneous multi-agent systems, an advisor may encounter tasks outside its competence yet still produce a fluent and confident answer~\citep{zhou2024larger, xiong2024can, sharma2024towards, kalai2025language}. To evaluate robustness to such misleading information provided by the advisors, we construct controlled corruptions by replacing a specified fraction of advisor responses with misleading ones. These responses remain fluent, on-topic, and well-formed, but their final answers are verified to be incorrect. We vary the misleading information ratio from $0\%$ to $100\%$~\footnote{a few advisors may still answer correctly at 100\% misleading due to limited sampling} to examine \textit{how consultation degrades as external evidence becomes less reliable}, and \textit{whether BaRe-Mem adaptively shifts between consultation and autonomous reasoning}.
\begin{figure}[!h]
\vspace{-1em}
\centering
\includegraphics[width=0.8\linewidth]{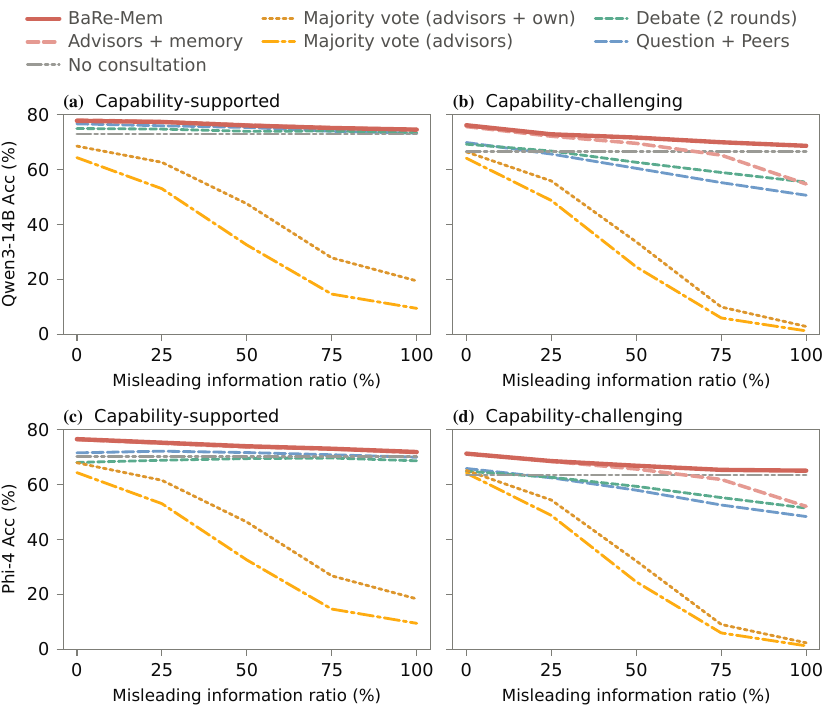}
\caption{
Accuracy under increasing misleading-advice ratios for Qwen3-14B and Phi-4.
\textbf{Left:} capability-supported regime; \textbf{right:} capability-challenging regime.
Across both regimes, BaRe-Mem remains above the no-consultation baseline as misleading information increases, while other consultation methods degrade substantially in the capability-challenging regime.
}
\vspace{-1em}
\label{fig:consultation_boundary}
\end{figure}

\textbf{{Obs.1. Consultation robustness depends on the capability regime.}}
Figure\textcolor{blue}{\textbf{~\ref{fig:consultation_boundary}}} shows a clear contrast between the two capability regimes. In the \textit{\textbf{capability-supported}} regime, \textbf{Question + Peers} and \textbf{Debate (2 rounds)} \textit{remain relatively stable as the misleading information ratio increases}. In the \textit{\textbf{capability-challenging}} regime, however, \textbf{both} \textit{degrade substantially} and \textit{fall below} the \textbf{No consultation} baseline once the misleading information ratio exceeds $50\%$ for both central models. This contrast is consistent with prior findings that the benefits of multi-agent interaction depend strongly on task and model capabilities~\citep{smit2023should,kim2026capable}. \textbf{Majority voting} is \textit{substantially more vulnerable}: both variants deteriorate rapidly as misleading information becomes dominant. Incorporating the central model's own answer partially mitigates this degradation, but does not prevent it, consistent with prior observations that models can abandon correct judgments in favor of incorrect peer majorities~\citep{weng2025we,zhu2025conformity}.

\textbf{Obs. 2. Reliability consultation helps but still requires autonomous reasoning.}
\textbf{Advisors + memory} is an ablation of BaRe-Mem that retains reliability guided attention but removes the autonomous option, forcing the model to consult on every question. \textbf{Advisors + memory} \textit{substantially improves robustness} over \textbf{Question + Peers}, indicating that historical reliability estimates make consultation less sensitive to misleading advisor responses. In the \textit{\textbf{capability-supported}} regime, this is often sufficient to maintain stable performance. However, in the \textit{\textbf{capability-challenging}} regime, its accuracy eventually falls below the \textbf{No consultation} baseline as misleading information becomes dominant for both central models. This exposes a limitation of relative advisor weighting: it can determine whom to trust more, but \textit{\textbf{not whether the advisor pool is worth consulting as a whole}}. \textbf{BaRe-Mem} \textit{adds this missing gap} by comparing estimated consultation and autonomous abilities on each question, and \textit{remains above} the \textbf{No consultation} baseline across all tested misleading information ratios. These results show that our BaRe-Mem helps the central model not only estimate source reliability, but also choose to rely on itself when external evidence is collectively unreliable.

\begin{table*}[!h]
\centering
\vspace{-0.5em}
\caption{
Accuracy (\%) and consultation ratio under increasing misleading information ratios for Qwen3-14B and Phi-4.
\textbf{No consultation} denotes autonomous reasoning without peer information, while \textbf{Question + Peers} denotes consultation with peer responses without reliability memory.
}
\label{tab:consultation_boundary_14b_Phi-4}
\scriptsize
\setlength{\tabcolsep}{5pt}
\begin{tabular}{ll|ccccc|ccccc}
\toprule
& &
\multicolumn{5}{c|}{\textbf{Capability-supported}} &
\multicolumn{5}{c}{\textbf{Capability-challenging}} \\
\cmidrule(lr){3-7}
\cmidrule(lr){8-12}
\textbf{Model} & \textbf{Method}
& 0\% & 25\% & 50\% & 75\% & 100\%
& 0\% & 25\% & 50\% & 75\% & 100\% \\
\midrule

\multirow{4}{*}{\qwen Qwen3-14B}
& No consultation
& 72.9 & 72.9 & 72.9 & 72.9 & 72.9
& 66.5 & 66.5 & 66.5 & 66.5 & 66.5 \\

& Question + Peers
& 76.6 & 75.9 & 75.3 & 73.9 & 73.2
& 69.8 & 65.6 & 60.4 & 55.2 & 50.6 \\

& \textbf{BaRe-Mem}
& \textbf{77.7} & \textbf{77.3} & \textbf{76.0} & \textbf{75.1} & \textbf{74.5}
& \textbf{76.1} & \textbf{72.8} & \textbf{71.6} & \textbf{69.9} & \textbf{68.6} \\

& Consultation ratio
& 87\% & 86\% & 85\% & 85\% & 85\%
& 90\% & 83\% & 76\% & 61\% & 21\% \\

\midrule

\multirow{4}{*}{\microsoft Phi-4}
& No consultation
& 70.2 & 70.2 & 70.2 & 70.2 & 70.2
& 63.4 & 63.4 & 63.4 & 63.4 & 63.4 \\

& Question + Peers
& 71.5 & 72.1 & 71.6 & 70.8 & 69.9
& 65.8 & 62.4 & 57.9 & 52.5 & 48.3 \\

& \textbf{BaRe-Mem}
& \textbf{76.5} & \textbf{75.2} & \textbf{73.9} & \textbf{73.0} & \textbf{71.8}
& \textbf{71.2} & \textbf{68.5} & \textbf{66.8} & \textbf{65.3} & \textbf{65.0} \\

& Consultation ratio
& 85\% & 84\% & 83\% & 84\% & 83\%
& 85\% & 79\% & 71\% & 57\% & 18\% \\

\bottomrule
\end{tabular}
\end{table*}

\textbf{Obs. 3. BaRe-Mem adapts when to consult.}
Table\textcolor{blue}{\textbf{~\ref{tab:consultation_boundary_14b_Phi-4}}}
reports the ratio of questions for which BaRe-Mem selects consultation.
In the \textit{\textbf{capability-supported}} regime, \textit{the consultation ratio remains high and nearly unchanged} as misleading information increases, decreasing only from $87\%$ to $85\%$ for Qwen3-14B and from $85\%$ to $83\%$ for Phi-4.
In contrast, in the \textit{\textbf{capability-challenging}} regime, it \textit{drops substantially} from $90\%$ to $21\%$ for Qwen3-14B and from $85\%$ to $18\%$ for Phi-4 as the misleading-information ratio increases from $0\%$ to $100\%$.
This behavior mirrors the performance patterns in Figure\textcolor{blue}{\textbf{~\ref{fig:consultation_boundary}}}: \textbf{BaRe-Mem} continues to consult when external information remains useful, but increasingly switches to autonomous reasoning as consultation becomes less reliable.

\subsection{Analysis of BaRe-Mem}
We next examine whether the quantities driving BaRe-Mem's consultation decisions behave as intended. Specifically, we study whether the memory captures the central model's autonomous ability, whether the predicted consultation advantage is consistent with the real empirical gain, and how much verified feedback is required to learn these estimates.
\subsubsection{Autonomous Ability Estimation}
\begin{figure}[!h]
\vspace{-0.5em}
\centering
\includegraphics[width=0.8\linewidth]{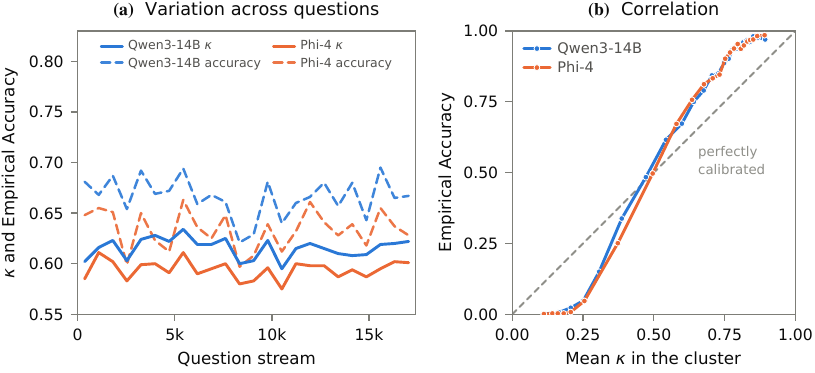}
\caption{
Estimated autonomous ability $\kappa$ versus empirical autonomous accuracy for Qwen3-14B and Phi-4.
\textbf{Left:} Evolution of mean $\kappa$ and empirical autonomous accuracy along the question stream.
\textbf{Right:} Empirical autonomous accuracy for groups of questions with similar $\kappa$ values. The dashed diagonal indicates perfect calibration.
}
\vspace{-0.8em}
\label{fig:kappa_analysis}
\end{figure}

BaRe-Mem decides whether to consult by comparing the estimated consultation ability with the central model's autonomous ability, making $\kappa_t$ a key quantity in the decision. We therefore examine whether reliability memory captures meaningful variation in the central model's own competence. As shown in Figure\textcolor{blue}{\textbf{~\ref{fig:kappa_analysis}}}, the estimated $\kappa$ \textbf{broadly tracks changes in empirical autonomous accuracy along the question stream} for both Qwen3-14B and Phi-4. To examine whether $\kappa$ is predictive of the central model's autonomous correctness, we group questions with similar $\kappa$ values and measure the autonomous accuracy within each group. \textbf{The empirical accuracy increases monotonically with $\kappa$}, indicating that higher estimated autonomous ability corresponds to a higher probability that the central model answers correctly on its own.

\subsubsection{Predicted vs. Real Consultation Gain}

\begin{figure}[!h]
\vspace{-1em}
\centering
\includegraphics[width=0.9\linewidth]{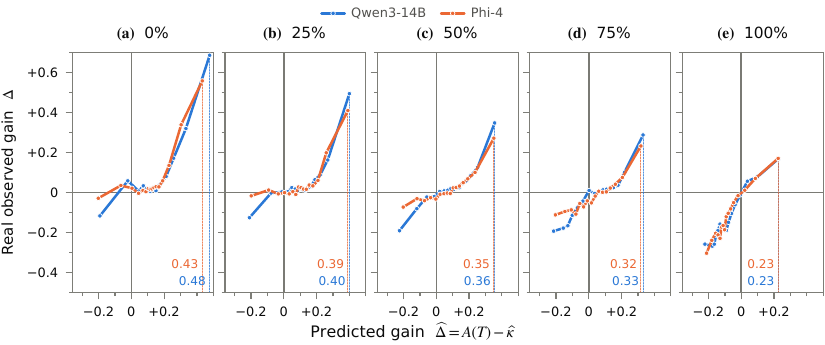}
\caption{
Predicted versus real consultation gain for Qwen3-14B and Phi-4 under increasing misleading-information ratios.
Panels (a)--(e) correspond to misleading-information ratios of $0\%$, $25\%$, $50\%$, $75\%$, and $100\%$, respectively.
The $x$-axis shows the mean predicted gain $\hat{\Delta}=A(T)-\kappa$ within each group, while the $y$-axis shows the corresponding mean real gain $\Delta=y^{\mathrm{Consult}}-y^{\mathrm{Direct}}$.
}
\vspace{-1em}
\label{fig:predicted_gain}
\end{figure}

We next examine whether the predicted gain of consultation over autonomous reasoning ($\hat{\Delta}_t=A(T_t)-\kappa_t$) is consistent with the real gain observed after verification. For each question $t$, we define the real gain as $\Delta_t=y_t^{\mathrm{Consult}}-y_t^{\mathrm{Direct}}\in\{-1,0,+1\}$, where $+1$ means consultation is correct while autonomous reasoning is wrong, $-1$ means the opposite, and $0$ means both modes have the same correctness outcome. We sort questions by $\hat{\Delta}_t$ and divide them into $16$ equal-sized groups. For each group, we plot the mean predicted gain on the $x$-axis and the mean real gain on the $y$-axis. The resulting curve approximates $\mathbb{E}[\Delta_t\mid\hat{\Delta}_t]$,
allowing us to examine whether the consultation advantage predicted by
BaRe-Mem is reflected in practice. Further analysis can be found in Appendix\textcolor{blue}{\textbf{~\ref{app:advisor_reliability_estimation}}}.

As shown in Figure\textcolor{blue}{\textbf{~\ref{fig:predicted_gain}}},
\textbf{the real gain increases with the predicted gain across all misleading information ratios}. Thus, when BaRe-Mem predicts a larger advantage from consultation, consultation is also more beneficial in practice. More importantly, \textbf{the curves cross zero close to $\hat{\Delta}=0$}, showing that the predicted boundary between consultation and autonomous reasoning is well aligned with the real boundary. As the misleading information ratio increases, the curves shift downward and to the left: consultation becomes less beneficial in practice, while \textbf{BaRe-Mem correspondingly predicts a smaller consultation advantage}. The upper range of $\hat{\Delta}$ also contracts as misleading information increases. Since $\kappa$ represents the central model's autonomous ability and is unaffected by external misleading information, this shift is primarily driven by a lower estimated consultation ability $A(T)$.

\vspace{-0.5em}
\subsubsection{Learning from Sparse Feedback}
\vspace{-0.5em}
\begin{figure}[h!]
\centering
\includegraphics[width=0.8\linewidth]{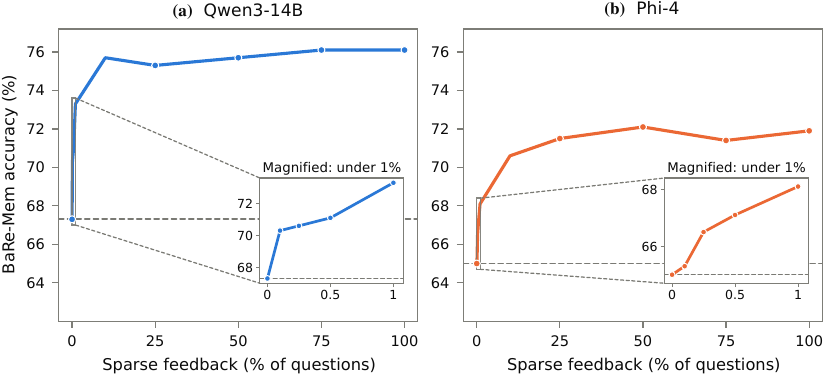}
\caption{
BaRe-Mem accuracy under sparse verified feedback for Qwen3-14B and Phi-4.
The fraction of questions whose verified outcomes are written to memory varies from $0\%$ to $100\%$.
Insets magnify the low-feedback regime below $1\%$.
}
\vspace{-1em}
\label{fig:sparse_feedback}
\end{figure}

In realistic deployments, verified outcomes may be costly or only intermittently available. We therefore examine how efficiently BaRe-Mem learns reliability from sparse feedback by varying the fraction of questions whose verified outcomes are written to memory. As shown in Figure\textcolor{blue}{\textbf{~\ref{fig:sparse_feedback}}}, both Qwen3-14B and Phi-4 \textbf{benefit substantially from even a small amount of verified feedback}. The magnified region below $1\%$ (approximately 170 samples) shows clear gains over the no-feedback setting, indicating that useful reliability estimates emerge after only a small number of verified interactions. Performance improves rapidly as feedback becomes available and then largely saturates, with most of the eventual gain obtained well before full feedback is provided. These results show that BaRe-Mem can learn useful reliability information without requiring dense verification.

\vspace{-0.5em}
\subsection{Real-World Application: Agent Team Routing}
\label{real-world_application}
We further evaluate BaRe-Mem in a realistic multi-agent team setting. A lead agent decomposes each task into sub-tasks, routes each sub-task to a worker, and replans by trying another worker when the returned report is rejected. The pipeline is shown in Figure \textcolor{blue}{\textbf{~\ref{fig:agent-team-pipeline}}}. We conduct this evaluation on MuSiQue~\citep{trivedi2022musique}, which contains 2,417 tasks and 6,404 sub-tasks. In this setting, BaRe-Mem is no longer used to reweight a fixed set of candidate responses. Instead, it is used to decide \textit{which worker should receive each sub-task} and \textit{which worker should be tried at first}. 

\begin{figure}[!h]
\centering
\vspace{-0.8em}
\includegraphics[width=0.8\linewidth]{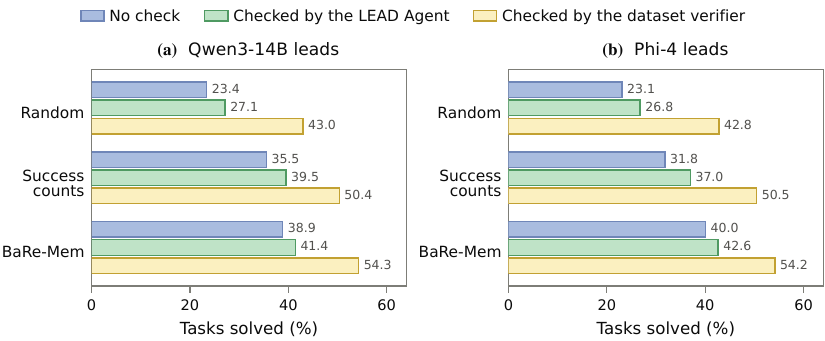}
\caption{
Agent-team task completion with Qwen3-14B and Phi-4 as lead agents.
We compare random routing, routing by historical success counts, and BaRe-Mem under three verification settings: no verification, verification by the lead agent, and exact verification by the dataset evaluator.
}
\vspace{-0.8em}
\label{fig:agent_team}
\end{figure}

Figure\textcolor{blue}{\textbf{~\ref{fig:agent_team}}} shows that \textbf{BaRe-Mem consistently achieves the highest task completion across both lead agents and all verification settings}, outperforming random routing\footnote{Random routing denotes the unmodified setting in which the lead agent selects workers on its own, without other intervention. It is not completely random allocation.} and routing by historical success counts. \textbf{Verification further improves all routing strategies}: lead agent checking provides a substantial gain over no verification, while exact dataset verification yields the highest completion rates. Importantly, the advantage of BaRe-Mem persists across verification settings, showing that its benefit comes from more effective worker routing rather than a particular verifier.

\begin{figure}[!h]
\centering
\includegraphics[width=0.8\linewidth]{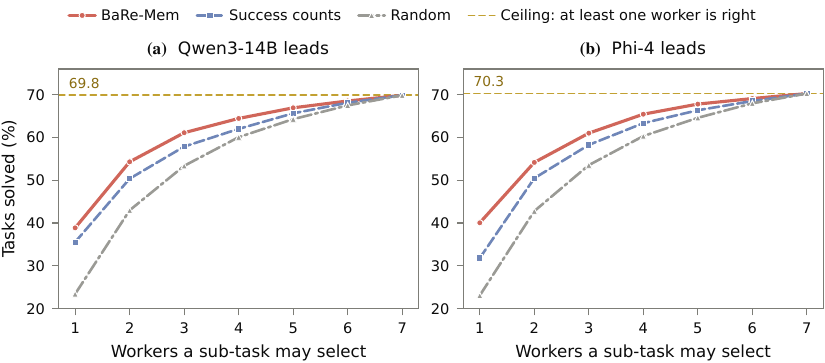}
\caption{
Task completion as the lead agent is allowed to try an increasing number of workers for each sub-task.
Subfigures (a) and (b) use Qwen3-14B and Phi-4 as the lead agent, respectively.
The dashed horizontal line denotes the ceiling at which at least one worker can solve the sub-task.
}
\label{fig:agent_team_sweep}
\end{figure}

To further understand why BaRe-Mem improves agent-team routing, we examine how quickly the lead agent can identify a worker capable of solving each sub-task. In this experiment, the dataset verifier checks the returned report after every worker call. If the checker returns a failure signal, the lead agent proceeds to the next ranked worker until the sub-task is solved or all candidates are exhausted. As shown in Figure~\textcolor{blue}{\textbf{\ref{fig:agent_team_sweep}}}, all routing strategies approach the same ceiling once every worker has been tried. However, \textbf{BaRe-Mem reaches higher task completion with fewer worker calls}, showing that its reliability memory helps the lead agent identify capable workers earlier in the search process. These results demonstrate that BaRe-Mem extends naturally beyond response attention to practical worker routing in realistic multi-agent systems.
\section{Conclusion}

We introduced \textbf{BaRe-Mem}, a Bayesian reliability memory that learns contextual source reliability from verified interaction history. By using reliability to both modulate advisor influence and decide whether consultation is preferable to autonomous reasoning, BaRe-Mem enables a central agent to adaptively determine \textit{whom to trust} and \textit{whether to consult}. Across heterogeneous tasks, it remains robust as advisor information becomes increasingly misleading, learns useful reliability from sparse feedback, and extends naturally to worker routing in agent teams. These results suggest that persistent reliability memory can provide a practical foundation for robust and adaptive consultation in long horizon multi-agent systems.

\bibliography{iclr2027_conference}
\bibliographystyle{iclr2027_conference}

\newpage

\appendix

\section{Mathematical Derivations}

\subsection{Bayesian Regression Posterior}
\label{apps:lemma1}

\begin{mdframed}[
  hidealllines=true,
  backgroundcolor=shadecolor,
  innerleftmargin=6pt,
  innerrightmargin=6pt,
  innertopmargin=4pt,
  innerbottommargin=4pt,
  skipabove=8pt,
  skipbelow=8pt
]
\noindent
Let $u_i\in\mathbb R^d$ be fixed inputs and consider
\[
z_i=u_i^\top\theta+\varepsilon_i,
\qquad
\varepsilon_i\sim\mathcal N(0,1),
\]
where the noises are mutually independent and independent of $\theta\sim\mathcal N(\theta_0,\Lambda_0^{-1}),$
with $\Lambda_0\succ0$.
For
$\mathcal D_n=\{(u_i,z_i)\}_{i=1}^n$,
the posterior is
\[
\theta\mid\mathcal D_n
\sim
\mathcal N(m_n,P_n^{-1}),
\]
where
\[
P_n
=
\Lambda_0+\sum_{i=1}^n u_i u_i^\top,
\qquad
q_n
=
\Lambda_0\theta_0+\sum_{i=1}^n u_i z_i,
\qquad
m_n=P_n^{-1}q_n.
\]
Moreover, $m_n$ uniquely minimizes
\[
J(\theta)
=
\sum_{i=1}^n(z_i-u_i^\top\theta)^2
+
\|\theta-\theta_0\|_{\Lambda_0}^2,
\]
\end{mdframed}

\noindent
\textbf{\textit{Proof.}}

\textbf{Part (i): \textit{Posterior derivation}}

For each observation, the regression model implies
$z_i\mid u_i,\theta\sim\mathcal N(u_i^\top\theta,1)$.
Therefore,
\begin{equation}
\begin{aligned}
p(z_i\mid u_i,\theta)
&=
\frac{1}{\sqrt{2\pi}}
\exp\!\left[
-\frac12(z_i-u_i^\top\theta)^2
\right],
\\
p(\mathcal D_n\mid\theta)
&=
\prod_{i=1}^n p(z_i\mid u_i,\theta)\propto
\exp\!\left[
-\frac12
\sum_{i=1}^n
(z_i-u_i^\top\theta)^2
\right],
\end{aligned}
\label{eq:likelihood}
\end{equation}
where the product factorization follows from the independence of the observation noises.

The Gaussian prior
$\theta\sim\mathcal N(\theta_0,\Lambda_0^{-1})$
has precision matrix $\Lambda_0$, and hence
\begin{equation}
p(\theta)
\propto
\exp\!\left[
-\frac12
(\theta-\theta_0)^\top
\Lambda_0
(\theta-\theta_0)
\right].
\label{eq:prior}
\end{equation}

Based on Bayes' rule:
$p(\theta\mid\mathcal D_n)
=
\frac{
p(\mathcal D_n\mid\theta)\,p(\theta)
}{
p(\mathcal D_n)
}$, we have
$p(\theta\mid\mathcal D_n)
\propto
p(\mathcal D_n\mid\theta)p(\theta)$.
Substituting Eqs. \textcolor{blue}{~(\ref{eq:likelihood})} and \textcolor{blue}{(\ref{eq:prior})},
taking $-2\log$, and collecting all terms independent of $\theta$
into $\mathrm{const}$ gives
\begin{equation}
\begin{aligned}
-2\log p(\theta\mid\mathcal D_n)
&=
\sum_{i=1}^n
(z_i-u_i^\top\theta)^2 +(\theta-\theta_0)^\top
\Lambda_0
(\theta-\theta_0)
+
\mathrm{const}.
\end{aligned}
\label{eq:posterior_negative_log}
\end{equation}

We next expand the two quadratic terms. For the likelihood term,
\begin{equation}
\begin{aligned}
\sum_{i=1}^n
(z_i-u_i^\top\theta)^2
&=
\sum_{i=1}^n
\left(
z_i^2
-
2z_i u_i^\top\theta
+
\theta^\top u_i u_i^\top\theta
\right)
\\
&=
\theta^\top
\left(
\sum_{i=1}^n u_i u_i^\top
\right)
\theta
-
2\theta^\top
\left(
\sum_{i=1}^n u_i z_i
\right)
+
\mathrm{const}.
\end{aligned}
\label{eq:likelihood_expand}
\end{equation}
Similarly, the prior term becomes
\begin{equation}
\begin{aligned}
(\theta-\theta_0)^\top
\Lambda_0
(\theta-\theta_0)
&=
\theta^\top\Lambda_0\theta
-
2\theta^\top\Lambda_0\theta_0
+
\theta_0^\top\Lambda_0\theta_0
\\
&=
\theta^\top\Lambda_0\theta
-
2\theta^\top\Lambda_0\theta_0
+
\mathrm{const}.
\end{aligned}
\label{eq:prior_expand}
\end{equation}

Combining
Eqs.\textcolor{blue}{~(\ref{eq:posterior_negative_log})},\textcolor{blue}{(~\ref{eq:likelihood_expand})}, and
\textcolor{blue}{(\ref{eq:prior_expand})}, we obtain
\begin{equation}
\begin{aligned}
-2\log p(\theta\mid\mathcal D_n)
&=
\theta^\top
\left(
\sum_{i=1}^n u_i u_i^\top+\Lambda_0
\right)
\theta
-
2\theta^\top
\left(
\sum_{i=1}^n u_i z_i+\Lambda_0\theta_0
\right)
+
\mathrm{const}
\\
&=
\theta^\top P_n\theta
-
2\theta^\top q_n
+
\mathrm{const}.
\end{aligned}
\label{eq:posterior_quadratic}
\end{equation}
where
$P_n=\Lambda_0+\sum_{i=1}^n u_i u_i^\top$
and
$q_n=\Lambda_0\theta_0+\sum_{i=1}^n u_i z_i$.

Because $\Lambda_0\succ0$ and each
$u_i u_i^\top\succeq0$, we have $P_n\succ0$. Therefore
$P_n^{-1}$ exists. Let $m_n=P_n^{-1}q_n$,
substituting in Eq.\textcolor{blue}{~(\ref{eq:posterior_quadratic})} gives
\begin{equation}
\theta^\top P_n\theta
-
2\theta^\top q_n
=
(\theta-m_n)^\top
P_n
(\theta-m_n)
-
m_n^\top P_n m_n.
\label{eq:complete_square}
\end{equation}
The final term does is independent of $\theta$ and can therefore be
absorbed into the constant. Hence,
\[
p(\theta\mid\mathcal D_n)
\propto
\exp\!\left[
-\frac12
(\theta-m_n)^\top
P_n
(\theta-m_n)
\right],
\]
which is the kernel of a Gaussian distribution with mean $m_n$
and precision $P_n$. 

Therefore, $\theta\mid\mathcal D_n\sim\mathcal N(m_n,P_n^{-1})$.

\noindent
\textbf{Part (ii): \textit{Posterior mean is the unique minimizer.}}

Next, we verify that the posterior mean is exactly the unique minimizer of the corresponding regularized least-squares objective. Our object:
\[
J(\theta)
=
\sum_{i=1}^n
(z_i-u_i^\top\theta)^2
+
\|\theta-\theta_0\|_{\Lambda_0}^2.
\]
From the expansion from Eq.\textcolor{blue}{~(\ref{eq:posterior_quadratic})},
$J(\theta)
=
\theta^\top P_n\theta
-
2\theta^\top q_n
+
\mathrm{const}$.
Thus,
\begin{equation}
    \nabla_\theta J(\theta)=2P_n\theta-2q_n
\end{equation}
Setting the gradient to zero gives
$P_n\theta=q_n$, and hence
$\theta=P_n^{-1}q_n=m_n$.
Since
\begin{equation}
    \nabla_\theta^2J(\theta)=2P_n\succ0
\end{equation}
Therefore, $m_n=P_n^{-1}q_n$ is the unique minimizer of $J(\theta)$.

\subsection{Deriving the Kalman gain}
\label{app:kalman_gain}

A fixed step update treats every verified observation equally, regardless of how uncertain the current posterior is in the direction of the new input. We therefore introduce a data dependent gain that determines how strongly the new residual should update the posterior mean, and choose it to minimize the expected squared estimation error.

Let $m_n$ and $S=P_n^{-1}$ denote the current posterior mean
and covariance. For a new fixed input $u$, the observation is
$z=u^\top\theta+\varepsilon$, where
$\varepsilon\sim\mathcal N(0,1)$ is independent of
$(\theta,\mathcal D_n)$.
We consider updating the posterior mean using the prediction residual:
\begin{equation}
\widetilde m(k)
=
m_n+k\bigl(z-u^\top m_n\bigr),
\label{eq:online_update_form}
\end{equation}
where the gain $k\in\mathbb R^d$ depends on the existing
observations and the input $u$, but not on the new outcome $z$.
We choose $k$ to minimize the expected squared estimation error,
conditional on $\mathcal D_n$.

Let $e=\theta-m_n$ be the current estimation error and
$e(k)=\theta-\widetilde m(k)$ the error after the update.
Substituting the observation model into
Eq.\textcolor{blue}{~(\ref{eq:online_update_form})} gives
\begin{equation}
\begin{aligned}
e(k)
&=
\theta-m_n
-k\bigl(u^\top\theta+\varepsilon-u^\top m_n\bigr)
\\
&=
e-k\bigl(u^\top e+\varepsilon\bigr)
\\
&=
(I-ku^\top)e-k\varepsilon.
\end{aligned}
\label{eq:online_error}
\end{equation}

Since $m_n$ and $S$ are the posterior mean and covariance, we want the error to satisfy
$\mathbb E[e\mid\mathcal D_n]=0$ and
$\mathbb E[ee^\top\mid\mathcal D_n]=S$. Because 
the new noise $\varepsilon$ has unit variance and is independent of
$(\theta,\mathcal D_n)$, so
$\mathbb E[e\varepsilon\mid\mathcal D_n]=0$.
Thus, taking the conditional expectation of the outer product
in Eq.\textcolor{blue}{~(\ref{eq:online_error})} eliminates the cross terms:
\begin{equation}
\begin{aligned}
C(k)
&\equiv
\mathbb E\bigl[e(k)e(k)^\top\mid\mathcal D_n\bigr]
\\
&=
(I-ku^\top)S(I-uk^\top)+kk^\top
\\
&=
S-Suk^\top-ku^\top S
+(1+v)kk^\top,
\end{aligned}
\label{eq:online_error_covariance}
\end{equation}
where $v=u^\top Su\ge0$ is the current posterior variance
of $u^\top\theta$.

The squared error satisfies
$\|e(k)\|_2^2=\operatorname{tr}(e(k)e(k)^\top)$.
Therefore, the conditional mean squared error is
\begin{equation}
\begin{aligned}
R(k)
&\equiv
\mathbb E\bigl[\|e(k)\|_2^2\mid\mathcal D_n\bigr]
\\
&=
\operatorname{tr}C(k)
\\
&=
\operatorname{tr}S
-2k^\top Su
+(1+v)k^\top k.
\end{aligned}
\label{eq:online_mse}
\end{equation}
The last equality uses the symmetry of $S$ and the trace
identities
$\operatorname{tr}(Suk^\top)=k^\top Su$ and
$\operatorname{tr}(ku^\top S)=u^\top Sk=k^\top Su$.

Differentiating with respect to $k$ gives
$\nabla_k R(k)=-2Su+2(1+v)k$.
Setting this gradient to zero yields
$(1+v)k=Su$.
Since the Hessian $2(1+v)I$ is positive definite,
the unique minimizer is
\begin{equation}
g
=
\frac{Su}{1+v}
=
\frac{P_n^{-1}u}{1+u^\top P_n^{-1}u}.
\label{eq:online_kalman_gain}
\end{equation}
This is the Kalman gain for the new observation.
It minimizes the conditional mean squared error among
updates of the form in Eq.\textcolor{blue}{~(\ref{eq:online_update_form})}.

\subsection{Exact Online Update}
\label{app:exact_online_update}

We now show that the posterior can be updated after each new
observation without recomputing the solution. We first prove
the rank-one inverse identity used in the update.

\begin{mdframed}[
  hidealllines=true,
  backgroundcolor=shadecolor,
  innerleftmargin=6pt,
  innerrightmargin=6pt,
  innertopmargin=4pt,
  innerbottommargin=4pt,
  skipabove=8pt,
  skipbelow=8pt
]
\noindent
\textbf{Property: Rank-one inverse identity.}
Let $P\succ0$, $S=P^{-1}$, and $u\in\mathbb R^d$. Then
\begin{equation}
(P+uu^\top)^{-1}
=
S
-
\frac{Suu^\top S}
{1+u^\top Su}.
\label{eq:rank_one_inverse}
\end{equation}
\end{mdframed}

\noindent
\textbf{\textit{Proof.}}
Let $c=u^\top Su$. Since $S\succ0$, we have $c\ge0$, so
$1+c>0$. Multiplying the right hand side of
Eq.\textcolor{blue}{~(\ref{eq:rank_one_inverse})} by $P+uu^\top$ gives
\begin{equation}
\begin{aligned}
&(P+uu^\top)
\left(
S-\frac{Suu^\top S}{1+c}
\right)
\\
&=
PS
-\frac{PSuu^\top S}{1+c}
+uu^\top S
-\frac{uu^\top Suu^\top S}{1+c}
\\
&=
I
-\frac{uu^\top S}{1+c}
+uu^\top S
-\frac{c\,uu^\top S}{1+c}
\\
&=
I
+
uu^\top S
-
\frac{(1+c)uu^\top S}{1+c}
\\
&=
I.
\end{aligned}
\end{equation}
Therefore,
$S-\frac{Suu^\top S}{1+c}$
is the inverse of $P+uu^\top$, which proves
Eq.\textcolor{blue}{~(\ref{eq:rank_one_inverse})}.
\hfill$\square$

\paragraph{Posterior Covariance Update.}
Suppose a new observation $(u,z)$ is added after $n$ observations.
The batch posterior parameters become
$P_{n+1}=P_n+uu^\top$ and $q_{n+1}=q_n+uz$.
Let $S_n=P_n^{-1}$ and define the Kalman gain derived in
Appendix\textcolor{blue}{\textbf{~\ref{app:kalman_gain}}} as
\begin{equation}
g
=
\frac{S_nu}
{1+u^\top S_nu}.
\label{eq:exact_update_gain}
\end{equation}
Applying Eq.\textcolor{blue}{~(\ref{eq:rank_one_inverse})} with
$P=P_n$ gives
\begin{equation}
\begin{aligned}
S_{n+1}
&\equiv
P_{n+1}^{-1}
\\
&=
S_n
-
\frac{S_nuu^\top S_n}
{1+u^\top S_nu}
\\
&=
S_n-g(S_nu)^\top.
\end{aligned}
\label{eq:blr_cov_update}
\end{equation}
Thus, the posterior covariance is updated by a rank-one
correction rather than by recomputing a matrix inverse.

\paragraph{Posterior Mean Update.}
To update the posterior mean, we first note that
\begin{equation}
\begin{aligned}
S_{n+1}u
&=
S_nu
-
g\,u^\top S_nu
\\
&=
S_nu
-
g\,c
\\
&=
(1+c)g-cg
=
g,
\end{aligned}
\label{eq:updated_cov_times_u}
\end{equation}
where $c=u^\top S_nu$. Using $m_n=S_nq_n$, we also have
\begin{equation}
    S_{n+1}q_n = (S_n-gu^\top S_n)q_n = m_n-g\,u^\top m_n.
\end{equation}
Therefore,
\begin{equation}
\begin{aligned}
m_{n+1}
&=
S_{n+1}q_{n+1}
\\
&=
S_{n+1}(q_n+uz)
\\
&=
m_n-g\,u^\top m_n+gz
\\
&=
m_n+g\bigl(z-u^\top m_n\bigr).
\end{aligned}
\label{eq:blr_mean_update}
\end{equation}

Eqs.\textcolor{blue}{~(\ref{eq:blr_cov_update})} and
\textcolor{blue}{~(\ref{eq:blr_mean_update})} are therefore exactly the covariance
and mean of the batch posterior after adding $(u,z)$.
Starting from $m_0=\theta_0$ and
$S_0=\Lambda_0^{-1}$, applying these updates recursively recovers
the batch posterior after every observation in exact arithmetic.

For BaRe-Mem, setting
$u=x_{t,k}$,
$z=s_{t,k}$,
and
$S_n=\Lambda^{-1}$
recovers the online memory updates used in the main text. This is the key mathematical principle behind the effectiveness of our BaRe-Mem.

\subsection{Order Independence}
\label{apps:order_independence}

\begin{mdframed}[
  hidealllines=true,
  backgroundcolor=shadecolor,
  innerleftmargin=6pt,
  innerrightmargin=6pt,
  innertopmargin=4pt,
  innerbottommargin=4pt,
  skipabove=8pt,
  skipbelow=8pt
]
\noindent
\textbf{Property: Order independence.}
For any fixed set of verified observations
$\mathcal D_n=\{(u_i,z_i)\}_{i=1}^n$,
the posterior maintained by BaRe-Mem is invariant to the order in which these observations are processed.
That is, for any permutation $\pi$ of $\{1,\ldots,n\}$,
\[
p(\theta\mid \mathcal D_n^{(\pi)})
=
p(\theta\mid \mathcal D_n),
\]
where
$\mathcal D_n^{(\pi)}
=
\{(u_{\pi(i)},z_{\pi(i)})\}_{i=1}^n$.
\end{mdframed}

\noindent
\textbf{\textit{Proof.}}
Consider any permutation $\pi$ of the $n$ observations. Processing them in the order
$\pi(1),\ldots,\pi(n)$ gives
\[
P_n^{(\pi)}
=
\Lambda_0+\sum_{i=1}^n u_{\pi(i)}u_{\pi(i)}^\top
=
\Lambda_0+\sum_{i=1}^n u_i u_i^\top
=
P_n,
\]
and similarly,
\[
q_n^{(\pi)}
=
\Lambda_0\theta_0+\sum_{i=1}^n u_{\pi(i)}z_{\pi(i)}
=
\Lambda_0\theta_0+\sum_{i=1}^n u_i z_i
=
q_n.
\]
Therefore,
$m_n^{(\pi)}
=
(P_n^{(\pi)})^{-1}q_n^{(\pi)}
=
P_n^{-1}q_n
=
m_n$,
and the posterior covariance is unchanged:
$(P_n^{(\pi)})^{-1}=P_n^{-1}$.
Hence,
$p(\theta\mid\mathcal D_n^{(\pi)})
=
p(\theta\mid\mathcal D_n)$.
\hfill$\square$

This property makes BaRe-Mem insensitive to the arrival order of a fixed set of verified evidence from different sources.

\subsection{Deriving the Reliability Estimate}
\label{app:record_estimate}

Given all verified observations so far, the reliability memory maintains
the posterior
$w\mid\mathcal D\sim\mathcal N(m,\Lambda^{-1})$.
For a candidate with representation $x$, the linear prediction
$w^\top x$ is therefore Gaussian. By the linear transformation
property of a Gaussian random variable,
\begin{equation}
w^\top x \mid x,\mathcal D
\sim
\mathcal N(\mu,v),
\qquad
\mu=x^\top m,
\qquad
v=x^\top\Lambda^{-1}x.
\label{eq:predictive_linear_score}
\end{equation}
Here, $\mu$ is the posterior mean of the signed correctness score,
while $v$ measures the uncertainty induced by the posterior over $w$.

Under the regression model, the signed correctness further includes
independent unit observation noise,
$s=w^\top x+\varepsilon$ with
$\varepsilon\sim\mathcal N(0,1)$.
Since $w^\top x$ and $\varepsilon$ are independent Gaussian random
variables, their sum is also Gaussian:
\begin{equation}
s\mid x,\mathcal D
\sim
\mathcal N(\mu,1+v).
\label{eq:predictive_signed_correctness}
\end{equation}

We use the probability that the predictive signed correctness is positive
as the reliability estimate of the candidate:
$p(x)\equiv\Pr(s>0\mid x,\mathcal D)$.
Standardizing $s$ using Eq.\textcolor{blue}{~(\ref{eq:predictive_signed_correctness})}
gives
\begin{equation}
\begin{aligned}
p(x)
&=
\Pr(s>0\mid x,\mathcal D)
\\
&=
\Pr\left(
\frac{s-\mu}{\sqrt{1+v}}
>
-\frac{\mu}{\sqrt{1+v}}
\right)
\\
&=
1-
\Phi\left(
-\frac{\mu}{\sqrt{1+v}}
\right)
\\
&=
\Phi\left(
\frac{\mu}{\sqrt{1+v}}
\right),
\end{aligned}
\label{eq:reliability_estimate_derivation}
\end{equation}
where $\Phi$ is the cumulative distribution function of the standard
normal distribution and the last equality follows from
$\Phi(-a)=1-\Phi(a)$.

Applying this result to candidate $k$ on question $q_t$, with
$x=x_{t,k}$, yields
\begin{equation}
p_{t,k}
=
\Phi\!\left(
\frac{\mu_{t,k}}
{\sqrt{1+v_{t,k}}}
\right),
\qquad
\mu_{t,k}
=
x_{t,k}^\top m,
\qquad
v_{t,k}
=
x_{t,k}^\top\Lambda^{-1}x_{t,k},
\end{equation}
which is the reliability estimate used in the Sec.\textcolor{blue}{\textbf{~\ref{reliability_estimate_000}}}  in the main text.

Especially, before any verified evidence is observed, the posterior mean equals the
zero-mean prior, $m=0$. Hence $\mu_{t,k}=0$ for every candidate and
$p_{t,k}=\Phi(0)=\tfrac12$, giving all candidates neutral initial
reliability. More generally, for a fixed $\mu_{t,k}$, increasing
$v_{t,k}$ decreases the magnitude of
$\mu_{t,k}/\sqrt{1+v_{t,k}}$ and therefore moves the reliability
estimate toward $\tfrac12$.

\subsection{Deriving the Estimates of $\rho$ and $\delta$}
\label{app:consultation_parameter_estimation}

Recall that in our main text Sec.\textcolor{blue}{\textbf{~\ref{Estimating_ability}}}, the consultation ability is modeled as
\begin{equation}
    A(T_t)
=
T_t\rho
+
(1-T_t)(\kappa_t-\delta).
\end{equation}
After question $t$ is verified, let $y_t\in\{0,1\}$ indicate
whether the consultation produces the correct answer. We model
$y_t$ with unit Gaussian noise:
\[
y_t
=
A(T_t)+\varepsilon_t,
\qquad
\varepsilon_t\sim\mathcal N(0,1).
\]
Substituting the expression for $A(T_t)$ gives
\begin{equation}
\begin{aligned}
y_t
&=
T_t\rho
+
(1-T_t)\kappa_t
-
(1-T_t)\delta
+
\varepsilon_t.
\end{aligned}
\label{eq:consultation_observation}
\end{equation}

Since $T_t$ and $\kappa_t$ are available from the reliability
memory, we move the known autonomous term to the left and define
\begin{equation}
    z_t
\equiv
y_t-(1-T_t)\kappa_t,
\qquad
u_t
=
\begin{bmatrix}
T_t\\
T_t-1
\end{bmatrix},
\qquad
\theta
=
\begin{bmatrix}
\rho\\
\delta
\end{bmatrix}.
\end{equation}
Eq. \textcolor{blue}{~(\ref{eq:consultation_observation})} then becomes
\begin{equation}
z_t
=
u_t^\top\theta+\varepsilon_t,
\qquad
\varepsilon_t\sim\mathcal N(0,1).
\label{eq:consultation_regression}
\end{equation}
Thus, estimating $\rho$ and $\delta$ reduces to a two-dimensional
Bayesian linear regression as we have proved in appendix\textcolor{blue}{\textbf{~\ref{apps:lemma1}}}

We place the Gaussian prior
$\theta\sim\mathcal N(\theta_0,I)$.
Given all verified interactions before question $t$, the posterior
mean is equivalently the unique minimizer of
\begin{equation}
J(\theta)
=
\sum_{s<t}
\bigl(z_s-u_s^\top\theta\bigr)^2
+
\|\theta-\theta_0\|_2^2.
\label{eq:consultation_objective}
\end{equation}
Expanding the objective gives
\[
J(\theta)
=
\theta^\top
\left(
I+\sum_{s<t}u_su_s^\top
\right)
\theta
-
2\theta^\top
\left(
\theta_0+\sum_{s<t}u_sz_s
\right)
+
\mathrm{const}.
\]
Therefore,
\begin{equation}
\nabla_\theta J(\theta)
=
2
\left(
I+\sum_{s<t}u_su_s^\top
\right)
\theta
-
2
\left(
\theta_0+\sum_{s<t}u_sz_s
\right).
\label{eq:consultation_gradient}
\end{equation}
Setting the gradient to zero gives
\begin{equation}
P_c\hat\theta=q_c,
\qquad
P_c
=
I+\sum_{s<t}u_su_s^\top,
\qquad
q_c
=
\theta_0+\sum_{s<t}u_sz_s.
\label{eq:consultation_normal_equations}
\end{equation}
Since $P_c\succ0$, the minimizer is unique, and hence
\begin{equation}
\hat\theta
=
\begin{bmatrix}
\hat\rho\\
\hat\delta
\end{bmatrix}
=
P_c^{-1}q_c.
\label{eq:consultation_parameter_estimate}
\end{equation}

The two components $\rho$ and $\delta$ have a direct interpretation. \textbf{\textit{(i)}} When $T_s=1$,
we have $u_s=[1,0]^\top$ and $z_s=y_s$, so the observation
contributes only to the estimate of $\rho$. At this endpoint,
$\rho$ represents consultation accuracy when trustworthy evidence
is available. \textbf{\textit{(ii)}} When $T_s=0$, we have $u_s=[0,-1]^\top$ and
$z_s=y_s-\kappa_s$, so the observation contributes only to the
estimate of $\delta$. In this case, $\delta$ measures the
shortfall of consultation relative to autonomous ability under
unreliable evidence. \textbf{\textit{(iii)}} For $0<T_s<1$, the observation provides
evidence about both parameters.

\subsection{Break-Even Trust and Decision Regimes}
\label{app:break_even}

The decision rule compares the estimated consultation ability
$A(T)$ with the autonomous ability $\kappa$. Define their difference as
\begin{equation}
\begin{aligned}
G(T)
&\equiv
A(T)-\kappa
\\
&=
T\hat\rho
+
(1-T)(\kappa-\hat\delta)
-
\kappa
\\
&=
T(\hat\rho-\kappa)
-
(1-T)\hat\delta
\\
&=
(\hat\rho+\hat\delta-\kappa)T
-
\hat\delta.
\end{aligned}
\label{eq:consultation_advantage}
\end{equation}
The central model consults exactly when $G(T)\ge0$.

\paragraph{Break-even trust.}
We first consider the regime used in the main text,
$\hat\delta>0$ and $\hat\rho>\kappa$.
At the two endpoints,
\[
G(0)=-\hat\delta<0,
\qquad
G(1)=\hat\rho-\kappa>0.
\]
Moreover,
$\hat\rho+\hat\delta-\kappa>0$, so $G(T)$ is strictly increasing
in $T$. Hence, there is a unique threshold $T^*\in(0,1)$ at which
consultation and autonomous reasoning have equal estimated ability.
Setting $G(T^*)=0$ gives
\begin{equation}
T^*
=
\frac{\hat\delta}
{\hat\rho+\hat\delta-\kappa}.
\label{eq:break_even_threshold}
\end{equation}
Therefore,
\[
A(T)\ge\kappa
\quad\Longleftrightarrow\quad
T\ge T^*.
\]
In this regime, consultation becomes preferable only when
the reliability of the available external evidence exceeds the
break-even trust $T^*$.

\paragraph{Dependence of the threshold.}
Let
$D=\hat\rho+\hat\delta-\kappa>0$.
Differentiating Eq.\textcolor{blue}{~(\ref{eq:break_even_threshold})} gives
\begin{equation}
\frac{\partial T^*}{\partial\kappa}
=
\frac{\hat\delta}{D^2}
>0,
\qquad
\frac{\partial T^*}{\partial\hat\delta}
=
\frac{\hat\rho-\kappa}{D^2}
>0,
\qquad
\frac{\partial T^*}{\partial\hat\rho}
=
-\frac{\hat\delta}{D^2}
<0.
\label{eq:break_even_derivatives}
\end{equation}
Hence, the break-even trust increases when the central model is
more capable autonomously or when unreliable consultation causes
greater degradation. In contrast, it decreases when the model is
better able to exploit trustworthy external evidence. A stronger
autonomous model therefore requires more reliable external evidence
before consultation becomes preferable.

\paragraph{Other parameter regimes.}
Because $G(T)$ is the linear function concerning $T$, its sign over $[0,1]$ is fully
determined by its endpoint values
$G(0)=-\hat\delta$ and
$G(1)=\hat\rho-\kappa$.
The remaining cases are summarized below:
\begin{table}[h]
\centering
\begin{tabular}{lll}
\hline
Condition
&
Behavior of $G(T)$
&
Decision
\\
\hline
$\hat\delta>0,\ \hat\rho\ge\kappa$
&
increases across zero
&
consult iff $T\ge T^*$
\\
$\hat\delta\ge0,\ \hat\rho<\kappa$
&
non-positive throughout
&
never consult
\\
$\hat\delta\le0,\ \hat\rho\ge\kappa$
&
non-negative throughout
&
always consult
\\
$\hat\delta<0,\ \hat\rho<\kappa$
&
decreases across zero
&
consult iff $T\le T^*$
\\
\hline
\end{tabular}
\end{table}

\subsection{Accuracy of Adaptive Consultation Selection}

Let $a_t$ and $o_t$ denote the probabilities that consultation and autonomous reasoning are correct on question $t$, respectively. Let $W$ be the set of questions on which the adaptive selection rule chooses the option with lower true accuracy.

For each question, the selected option is correct with probability
\begin{equation}
c_t
=
\begin{cases}
\max(a_t,o_t), & t\notin W,\\
\min(a_t,o_t)
=
\max(a_t,o_t)-|a_t-o_t|, & t\in W.
\end{cases}
\end{equation}

By linearity of expectation, the expected accuracy over $N$ questions is
\begin{equation}
\mathrm{Acc}_{\mathrm{select}}
=
\frac{1}{N}\sum_{t=1}^N \max(a_t,o_t)
-
\frac{1}{N}\sum_{t\in W}|a_t-o_t|.
\end{equation}

Since $\max(a_t,o_t)\ge a_t$ and $\max(a_t,o_t)\ge o_t$ for every $t$,
\begin{equation}
\mathrm{Acc}_{\mathrm{select}}
\ge
\max\!\left(
\mathrm{Acc}_{\mathrm{consult}},
\mathrm{Acc}_{\mathrm{alone}}
\right)
-
\frac{1}{N}\sum_{t\in W}|a_t-o_t|.
\end{equation}

Therefore, when the selection rule makes no wrong-side decisions
($W=\varnothing$), adaptive selection is at least as accurate as always consulting or always reasoning autonomously.

This characterization also clarifies what the decision module needs to estimate accurately. The final choice depends on the ordering between consultation and autonomous reasoning, rather than on perfectly calibrated absolute ability estimates. In particular, the decision is correct whenever the estimated quantities preserve the sign of $A(T_t)-\kappa_t$.
When this ordering is reversed, the resulting penalty is exactly the true accuracy gap $|a_t-o_t|$. Thus, errors are most consequential on questions where consultation and autonomous reasoning differ substantially, while miscalibration near ties has little effect on final accuracy.

\section{Additional Experiments and Analysis}
\label{app:additional_experiments}
\subsection{Capability of Advisors and Central Models}
\label{app:model_performance}
Tables\textcolor{blue}{\textbf{~\ref{tab:capability-supported-models}}} and\textcolor{blue}{\textbf{~\ref{tab:capability-challenging-models}}} report the individual performance of all advisor and central models used in our experiments. These results shows the capability structure underlying the two evaluation regimes. In the \textbf{\textit{capability-supported}} regime, multiple models exhibit strong competence on the corresponding tasks, and the advisor pool frequently contains at least one correct response. In the \textbf{\textit{capability-challenging}} regime, performance varies more substantially across both tasks and models, producing stronger task-dependent heterogeneity. This variation provides the setting in which contextual reliability estimation is important.
\begin{table}[!h]
\centering
\caption{
Accuracy (\%) of the six advisor models and six central models in the capability-supported regime.
Numbers below each dataset name indicate the number of evaluation examples.
\textbf{At least one right} reports the fraction of examples for which at least one advisor provides a correct answer.
}
\label{tab:capability-supported-models}

\setlength{\tabcolsep}{7pt}
\renewcommand{\arraystretch}{0.95}

\begin{tabular}{lccc}
\toprule
Capability-supported
& \begin{tabular}[c]{@{}c@{}}GSM8K\\ 1,319\end{tabular}
& \begin{tabular}[c]{@{}c@{}}SQuAD\\ 2,000\end{tabular}
& \begin{tabular}[c]{@{}c@{}}APPS\\ 1,000\end{tabular} \\
\midrule

\multicolumn{4}{c}{Advisor} \\
\midrule

\google Gemma-3-4B
& 67.4\% & 17.4\% & 16.8\% \\

\microsoft Phi-4-mini
& 56.9\% & 85.9\% & 19.4\% \\

\qwen Qwen2.5-Coder-7B
& 53.6\% & 75.8\% & 17.3\% \\

\meta Llama-3.1-8B
& 81.3\% & 68.7\% & 12.1\% \\

\deepseek DeepSeek-Coder-V2-Lite
& 81.4\% & 20.1\% & 26.6\% \\

\deepseek R1-Distill-Qwen-7B
& 87.3\% & 29.8\% & 23.1\% \\

\midrule
At least one right
& 96.1\% & 92.7\% & 43.2\% \\

\midrule
\multicolumn{4}{c}{Central Model} \\
\midrule

\qwen Qwen3-4B
& 89.2\% & 77.7\% & 25.8\% \\

\qwen Qwen3-8B
& 91.9\% & 82.6\% & 28.5\% \\

\qwen Qwen3-14B
& 94.2\% & 78.7\% & 33.4\% \\

\ministra Ministral-8B
& 64.6\% & 72.8\% & 13.1\% \\

\qwen Qwen2.5-7B
& 91.3\% & 83.8\% & 16.8\% \\

\microsoft Phi-4
& 94.4\% & 70.2\% & 38.3\% \\

\bottomrule
\end{tabular}
\end{table}

The \textbf{At least one right} row measures the oracle coverage of the advisor pool. High coverage indicates that useful external evidence is often available even when no single advisor is consistently reliable across tasks~\citep{shen2023hugginggpt, jiang2023llm,wang2025mixture}. At the same time, the large variation in individual model accuracy shows why source identity alone is insufficient: the most reliable advisor depends strongly on the task under consideration.

\begin{table}[!h]
\centering
\caption{
Accuracy (\%) of the six advisor models and six central models in the capability-challenging regime.
Numbers below each dataset name indicate the number of evaluation examples.
\textbf{At least one right} reports the fraction of examples for which at least one advisor produces a correct answer.
}
\label{tab:capability-challenging-models}

\setlength{\tabcolsep}{5pt}
\renewcommand{\arraystretch}{0.95}

\begin{tabular}{lcccccc}
\toprule
Capability-challenging
& \begin{tabular}[c]{@{}c@{}}PIQA\\ 1,838\end{tabular}
& \begin{tabular}[c]{@{}c@{}}MMLU\\ 1,531\end{tabular}
& \begin{tabular}[c]{@{}c@{}}OpenBookQA\\ 500\end{tabular}
& \begin{tabular}[c]{@{}c@{}}SciQ\\ 1,000\end{tabular}
& \begin{tabular}[c]{@{}c@{}}BBH\\ 6,511\end{tabular}
& \begin{tabular}[c]{@{}c@{}}SuperGLUE\\ 6,023\end{tabular} \\
\midrule

\multicolumn{7}{c}{Advisor} \\
\midrule

\google Gemma-3-4B
& 77.0\% & 54.1\% & 72.8\% & 87.1\% & 28.7\% & 77.5\% \\

\microsoft Phi-4-mini
& 80.1\% & 64.1\% & 76.4\% & 90.1\% & 10.7\% & 79.1\% \\

\qwen Qwen2.5-Coder-7B
& 84.4\% & 56.2\% & 79.8\% & 92.1\% & 12.8\% & 80.7\% \\

\meta Llama-3.1-8B
& 82.2\% & 61.2\% & 81.6\% & 93.9\% & 27.6\% & 77.5\% \\

\deepseek DeepSeek-Coder-V2-Lite
& 73.2\% & 23.4\% & 38.2\% & 32.7\% & 3.7\% & 65.1\% \\

\deepseek R1-Distill-Qwen-7B
& 56.4\% & 54.4\% & 66.2\% & 79.1\% & 45.6\% & 76.2\% \\

\midrule
At least one right
& 97.8\% & 90.1\% & 97.6\% & 98.5\% & 61.1\% & 96.3\% \\

\midrule
\multicolumn{7}{c}{Central Model} \\
\midrule

\qwen Qwen3-4B
& 82.6\% & 74.1\% & 84.0\% & 94.0\% & 41.9\% & 84.0\% \\

\qwen Qwen3-8B
& 87.3\% & 79.1\% & 88.2\% & 95.4\% & 34.5\% & 82.3\% \\

\qwen Qwen3-14B
& 89.1\% & 81.3\% & 88.8\% & 95.9\% & 33.1\% & 85.5\% \\

\ministra Ministral-8B
& 81.2\% & 67.7\% & 84.0\% & 93.3\% & 23.7\% & 79.4\% \\

\qwen Qwen2.5-7B
& 86.7\% & 74.6\% & 86.8\% & 94.2\% & 21.6\% & 79.0\% \\

\microsoft Phi-4
& 91.0\% & 84.9\% & 89.8\% & 96.3\% & 23.9\% & 84.4\% \\

\bottomrule
\end{tabular}
\end{table}

In the \textbf{\textit{capability-challenging}} regime, model performance varies substantially across tasks, confirming strong task-dependent heterogeneity among the advisors. Nevertheless, the \textbf{At least one right} score remains high on most datasets, indicating that useful external evidence is still provided in the advisor pool. The challenge is therefore not simply obtaining external information, but identifying which advisor provides reliable evidence for the current question and determining how strongly that evidence should influence the central model.

\subsection{Consultation Gains Across Central Models}
\begin{figure}[!h]
\centering
\includegraphics[width=1.0\linewidth]{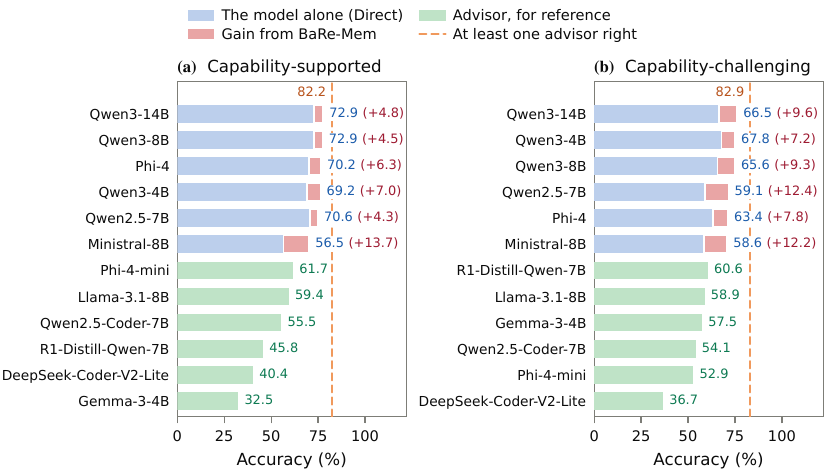}
\caption{
Direct accuracy and gains from BaRe-Mem across central models.
\textbf{Left:} capability-supported regime; \textbf{right:} capability-challenging regime.
Blue bars show autonomous accuracy, red segments show the additional gain from BaRe-Mem, and green bars show individual advisor accuracy for reference.
The dashed line denotes the fraction of questions for which at least one advisor is correct.
}
\vspace{-0.8em}
\label{fig:central_model_gain}
\end{figure}

Figure\textcolor{blue}{\textbf{~\ref{fig:central_model_gain}}} shows that \textbf{BaRe-Mem improves every evaluated central model in both capability regimes}. The gains are particularly pronounced in the \textbf{\textit{capability-challenging}} regime, where model and advisor capabilities are more heterogeneous. Improvements range from $7.2$ to $12.4$ accuracy points across the six central models. Even in the \textbf{\textit{capability-supported}} regime, where the central models are already relatively strong, BaRe-Mem consistently provides additional gains of $4.3$--$13.7$ points.

\subsection{Advisor Selection Behavior}
\begin{figure}[!h]
\centering
\includegraphics[width=1.0\linewidth]{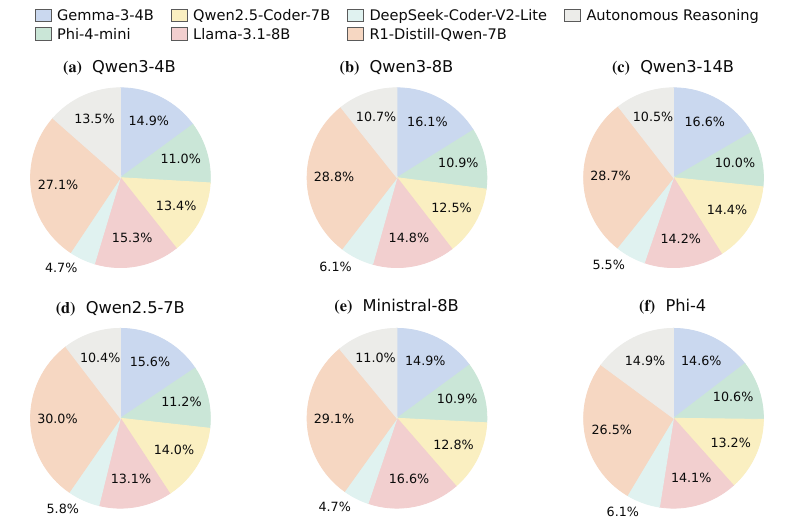}
\caption{
Distribution of autonomous reasoning and advisor attribution across central models under 0\% misleading advice ratio.
Autonomous reasoning denotes questions answered without consultation; all remaining questions are attributed to the advisor with the highest reliability estimate before feedback on the current question.
Advisor shares therefore reflect reliability based ranking rather than exclusive use of a single advisor's response.
}
\vspace{-0.8em}
\label{fig:advisor_selection}
\end{figure}

Figure\textcolor{blue}{\textbf{~\ref{fig:advisor_selection}}}
shows two consistent patterns in BaRe-Mem's consultation behavior.
First, \deepseek( \qwen) \textbf{R1-Distill-Qwen-7B} \textit{receives the largest advisor attribution}
across all six central models, while \deepseek \textbf{DeepSeek-Coder-V2-Lite} \textit{receives
the smallest}, indicating a broadly consistent reliability ordering
under uncorrupted advice.
However, \textbf{the attribution shares vary across central models}, showing
that the learned reliability is not determined by source identity alone.
Second, \textbf{BaRe-Mem retains autonomous reasoning for every central model}
rather than forcing consultation on all questions.

\newpage
\subsection{Advisor Reliability Estimation}
\label{app:advisor_reliability_estimation}
\begin{figure}[!h]
\centering
\includegraphics[width=1.0\linewidth]{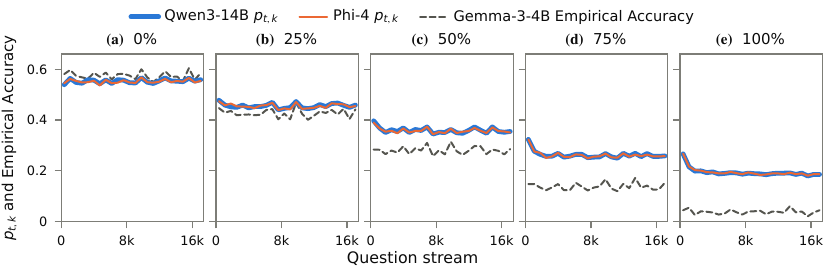}
\caption{
Online reliability estimation for advisor Gemma-3-4B.
Subfigures (a)--(e) correspond to misleading information ratios of $0\%$, $25\%$, $50\%$, $75\%$, and $100\%$, respectively.
Solid curves show the pre-feedback reliability estimate $p_{t,k}$ produced by Qwen3-14B and Phi-4, while the dashed curve shows the empirical accuracy of Gemma-3-4B.
}
\vspace{-0.8em}
\label{fig:advisor_reliability_gemma}
\end{figure}

\begin{figure}[!h]
\centering
\includegraphics[width=1.0\linewidth]{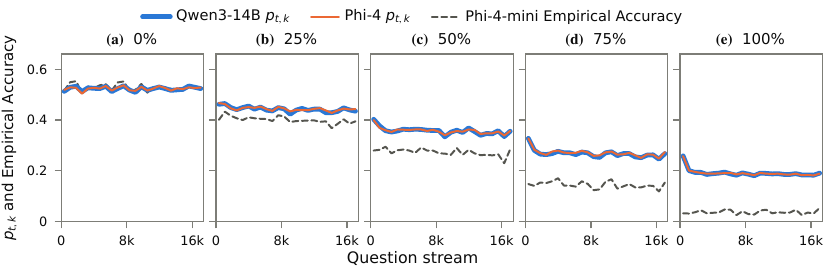}
\caption{
Online reliability estimation for advisor Phi-4-mini.
Subfigures (a)--(e) correspond to misleading information ratios of $0\%$, $25\%$, $50\%$, $75\%$, and $100\%$, respectively.
Solid curves show the pre-feedback reliability estimate $p_{t,k}$ produced by Qwen3-14B and Phi-4, while the dashed curve shows the empirical accuracy of Phi-4-mini.
}
\vspace{-0.8em}
\label{fig:advisor_reliability_phi4mini}
\end{figure}

\begin{figure}[!h]
\centering
\includegraphics[width=1.0\linewidth]{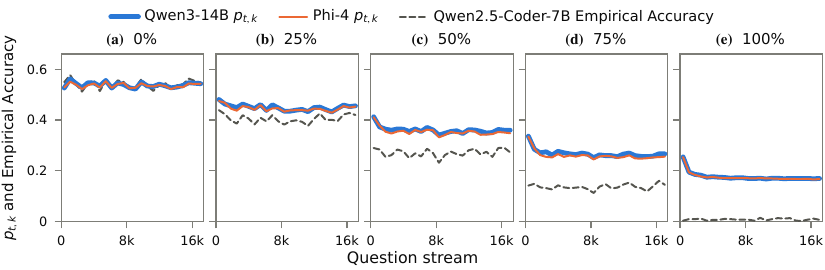}
\caption{
Online reliability estimation for advisor Qwen2.5-Coder-7B.
Subfigures (a)--(e) correspond to misleading information ratios of $0\%$, $25\%$, $50\%$, $75\%$, and $100\%$, respectively.
Solid curves show the pre-feedback reliability estimate $p_{t,k}$ produced by Qwen3-14B and Phi-4, while the dashed curve shows the empirical accuracy of Qwen2.5-Coder-7B.
}
\vspace{-0.8em}
\label{fig:advisor_reliability_qwencoder}
\end{figure}

\begin{figure}[!h]
\centering
\includegraphics[width=1.0\linewidth]{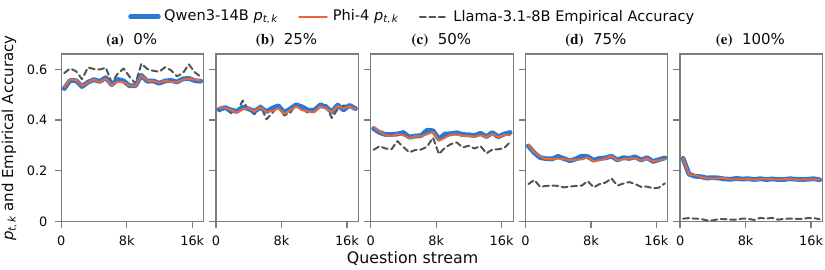}
\caption{
Online reliability estimation for advisor Llama-3.1-8B.
Subfigures (a)--(e) correspond to misleading information ratios of $0\%$, $25\%$, $50\%$, $75\%$, and $100\%$, respectively.
Solid curves show the pre-feedback reliability estimate $p_{t,k}$ produced by Qwen3-14B and Phi-4, while the dashed curve shows the empirical accuracy of Llama-3.1-8B.
}
\vspace{-0.8em}
\label{fig:advisor_reliability_llama}
\end{figure}

\begin{figure}[!h]
\centering
\includegraphics[width=1.0\linewidth]{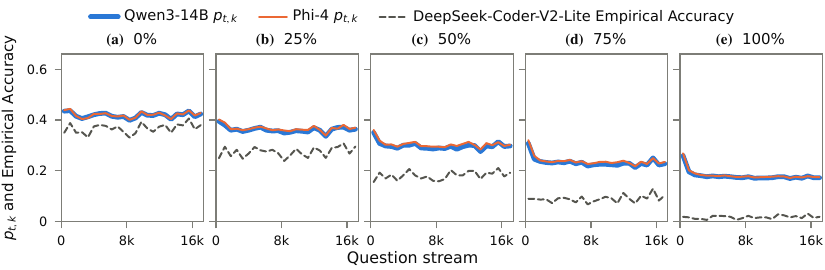}
\caption{
Online reliability estimation for advisor DeepSeek-Coder-V2-Lite.
Subfigures (a)--(e) correspond to misleading information ratios of $0\%$, $25\%$, $50\%$, $75\%$, and $100\%$, respectively.
Solid curves show the pre-feedback reliability estimate $p_{t,k}$ produced by Qwen3-14B and Phi-4, while the dashed curve shows the empirical accuracy of DeepSeek-Coder-V2-Lite.
}
\vspace{-0.8em}
\label{fig:advisor_reliability_dscoder}
\end{figure}

\begin{figure}[!h]
\centering
\includegraphics[width=1.0\linewidth]{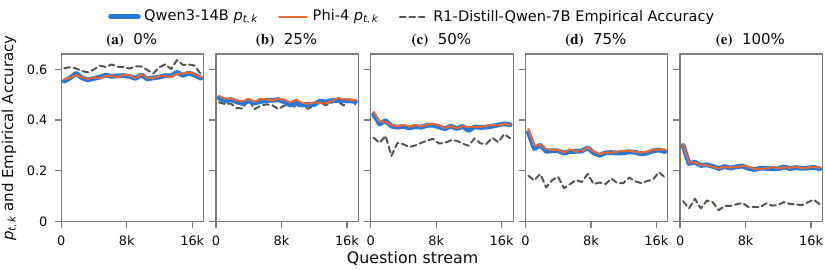}
\caption{
Online reliability estimation for advisor R1-Distill-Qwen-7B.
Subfigures (a)--(e) correspond to misleading information ratios of $0\%$, $25\%$, $50\%$, $75\%$, and $100\%$, respectively.
Solid curves show the pre-feedback reliability estimate $p_{t,k}$ produced by Qwen3-14B and Phi-4, while the dashed curve shows the empirical accuracy of R1-Distill-Qwen-7B.
}
\vspace{-0.8em}
\label{fig:advisor_reliability_r1distill}
\end{figure}

Figures\textcolor{blue}{\textbf{~\ref{fig:advisor_reliability_gemma}--\ref{fig:advisor_reliability_r1distill}}}
examine whether the reliability memory estimation reflects advisor performance
under increasingly misleading information.
Across all six advisors, higher misleading information ratios lead
to lower empirical accuracy and lower reliability estimates from
both Qwen3-14B and Phi-4.
The estimates therefore reflect the deterioration of external
evidence rather than remaining stable.

Within the high misleading information regime, \textbf{the estimates generally decline
more rapidly near the beginning and change less later}.
This behavior is consistent with the Kalman gain update derived in
Appendix\textcolor{blue}{\textbf{~\ref{app:kalman_gain}}}.
For a fixed representation, the update corrects a fraction
$v/(1+v)$ of the prediction residual, where $v$ is the current
posterior variance in that direction.
As accumulated evidence reduces this uncertainty, subsequent
observations produce smaller corrections for the same residual.

However, \textbf{the estimates remain above empirical accuracy at high
misleading information ratios}. Especially, in the 100\% misleading information regime, the empirical accuracy is nearly 0, but the estimates remain around 0.2.
The Gaussian predictive mapping can explain this discrepancy.
With repeated incorrect outcomes at a fixed nonzero representation,
the regression mean approaches the signed target $-1$ and its
posterior variance approaches zero. Nevertheless,
\begin{equation}
p
=
\Phi\!\left(\frac{\mu}{\sqrt{1+v}}\right)
\longrightarrow
\Phi(-1)
\approx 0.159.
\label{eq:reliability_incorrect_limit}
\end{equation}
The unit observation noise remains even when uncertainty about
the regression parameters vanishes, leaving nonzero predictive
probability above zero.
This is an illustrative limit rather than a universal lower bound,
since the linear predictor is unconstrained.
The figures thus support the memory's responsiveness to verified
feedback, while also revealing a calibration gap between its
reliability estimates and empirical correctness probabilities.

\clearpage
\subsection{Consultation Robustness Across Additional Central Models}
\label{app:consultation_boundary}

\begin{figure}[!h]
\centering
\includegraphics[width=0.9\linewidth]{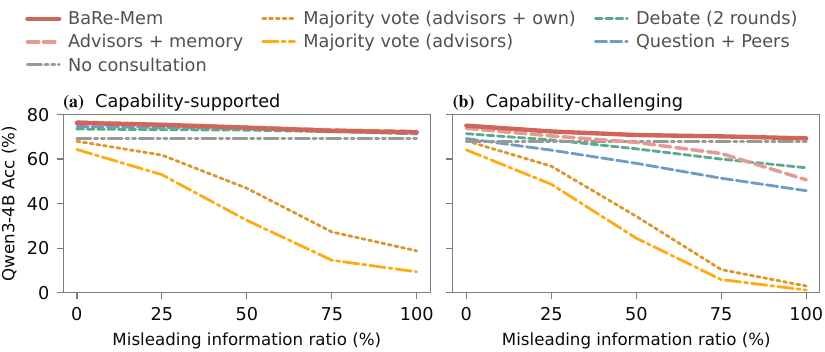}
\caption{
Accuracy of Qwen3-4B as the central model under increasing misleading information ratios.
\textbf{Left:} capability-supported regime; \textbf{right:} capability-challenging regime.
We compare BaRe-Mem with \textit{Advisors + memory}, \textit{No consultation}, \textit{Question + Peers}, \textit{Debate (2 rounds)}, and two majority voting baselines.
}
\vspace{-0.8em}
\label{fig:qwen3-4b-misleading}
\end{figure}

\begin{figure}[H]
\centering
\includegraphics[width=0.9\linewidth]{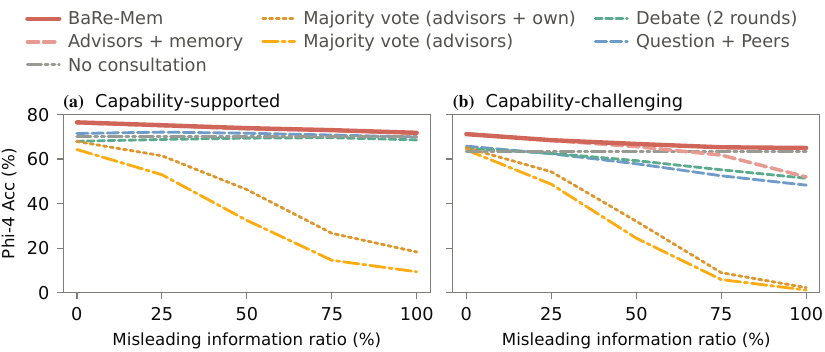}
\caption{
Accuracy of Phi-4 as the central model under increasing misleading information ratios.
\textbf{Left:} capability-supported regime; \textbf{right:} capability-challenging regime.
We compare BaRe-Mem with \textit{Advisors + memory}, \textit{No consultation}, \textit{Question + Peers}, \textit{Debate (2 rounds)}, and two majority voting baselines.
}
\vspace{-0.8em}
\label{fig:phi4-misleading}
\end{figure}

We report complete results for six central models: \qwen Qwen3-4B, \qwen Qwen3-8B, \qwen Qwen3-14B, \qwen Qwen2.5-7B, \ministra Ministral-8B, and \microsoft Phi-4. Tables~\textcolor{blue}{\textbf{\ref{tab:capability-supported-challenging}}}--\textcolor{blue}{\textbf{\ref{tab:phi-4-capability-comparison}}} and Figures~\textcolor{blue}{\textbf{\ref{fig:qwen3-4b-misleading}}}--\textcolor{blue}{\textbf{\ref{fig:phi4-misleading}}} extend the analysis in the main text Sec. ~\textcolor{blue}{\textbf{\ref{Adaptive Consultation under Misleading Information}}} across different model sizes and families.

\textbf{Obs. 1. Historical reliability improves consultation across models.}
Across both capability regimes and all tested misleading information ratios, \textbf{Advisors + memory} a\textit{chieves higher accuracy than} \textbf{Question + Peers} for every central model. At $50\%$ misleading information in the \textbf{\textit{capability-challenging}} regime, adding reliability memory improves accuracy by $9.4\%$, $12.6\%$, and $9.1\%$ percentage points for \qwen Qwen3-4B, Qwen3-8B, and Qwen3-14B, respectively. The corresponding improvements are $11.8\%$ points for \qwen Qwen2.5-7B, $19.9\%$ points for \ministra Ministral-8B, and $7.7\%$ points for \microsoft Phi-4. Since both methods consult on every question, this comparison demonstrates the benefit of incorporating historical reliability into the consultation process. The consistent gains support using accumulated correctness evidence to guide advisor influence rather than relying solely on the current responses.

\textbf{Obs. 2. Autonomous selection complements reliability-guided consultation.}
The benefit of choosing whether to consult becomes particularly clear when misleading information dominates. At $100\%$ misleading information in the \textbf{\textit{capability-challenging}} regime, \textbf{Question + Peers}, \textbf{Debate (2 rounds)}, and \textbf{both majority voting} baselines \textit{fall below} \textbf{No consultation} for all six central models. Although \textbf{Advisors + memory} improves consultation, it also \textit{falls below} the \textbf{autonomous} baseline under this condition. In contrast, \textbf{BaRe-Mem} \textit{exceeds its always consult ablation} by $13.0\%$--$34.6\%$ percentage points and remains $1.5\%$--$2.8\%$ points \textit{above autonomous reasoning}. Moreover, it \textit{stays above} the \textbf{autonomous baseline} throughout the tested misleading information range in this regime for all six models. These results support the complementary roles of the two mechanisms: reliability-guided attention improves how advisor responses are used, while autonomous selection provides an alternative when continued consultation becomes harmful.

\textbf{Obs. 3. Consultation behavior adapts to the model and capability regime.}
The consultation ratios provide a behavioral explanation for these results. In the \textbf{\textit{capability-supported}} regime, \qwen Qwen3-4B, Qwen3-8B, Qwen3-14B, and \microsoft Phi-4 maintain high consultation ratios, with endpoint decreases of only $1\%$--$3\%$ points as misleading information increases from $0\%$ to $100\%$. \qwen Qwen2.5-7B and \ministra Ministral-8B reduce consultation more noticeably, from $84\%$ to $77\%$ and from $85\%$ to $71\%$, respectively, but still consult on most questions. In the \textbf{\textit{capability-challenging}} regime, \textbf{all six models exhibit a much stronger shift}: consultation ratios decrease from $85\%$--$90\%$ without injected misleading information to $9\%$--$21\%$ at the highest misleading information ratio. Thus, the same increase in misleading information produces different consultation behavior across models and capability regimes. This pattern is consistent with comparing estimated consultation and autonomous abilities on each question, rather than applying a fixed consultation policy.

\begin{table}[!h]
\centering
\caption{
Accuracy (\%) of consultation methods with Qwen3-4B as the central model under increasing misleading information ratios.
Results are reported for the capability-supported and capability-challenging regimes.
Bold numbers indicate the best performance in each column.
}
\label{tab:capability-supported-challenging}
\resizebox{\textwidth}{!}{
\begin{tabular}{l|ccccc|ccccc}
\toprule
\multirow{2}{*}{\textbf{Method}}
& \multicolumn{5}{c|}{\textbf{Capability-supported}}
& \multicolumn{5}{c}{\textbf{Capability-challenging}} \\
\cmidrule(lr){2-6} \cmidrule(lr){7-11}
& \textbf{0\%} & \textbf{25\%} & \textbf{50\%} & \textbf{75\%} & \textbf{100\%}
& \textbf{0\%} & \textbf{25\%} & \textbf{50\%} & \textbf{75\%} & \textbf{100\%} \\
\midrule

No consultation
& 69.2 & 69.2 & 69.2 & 69.2 & 69.2
& 67.8 & 67.8 & 67.8 & 67.8 & 67.8 \\

Question + Peers
& 74.6 & 74.2 & 73.7 & 72.4 & 71.3
& 69.2 & 64.0 & 58.1 & 51.4 & 45.8 \\

Majority vote (advisors)
& 64.3 & 53.0 & 32.5 & 14.6 & 9.4
& 64.1 & 48.7 & 24.5 & 5.9 & 1.2 \\

Majority vote (advisors + own)
& 68.0 & 61.8 & 46.9 & 27.3 & 18.8
& 68.2 & 56.8 & 34.3 & 10.4 & 3.0 \\

Debate (2 rounds)
& 73.5 & 73.2 & 73.0 & 72.3 & 72.1
& 71.4 & 68.6 & 64.6 & 60.0 & 56.1 \\

Advisors + memory
& \textbf{76.7} & \textbf{75.5} & \textbf{74.3} & \textbf{73.1} & \textbf{72.4}
& 73.8 & 70.4 & 67.5 & 62.5 & 50.7 \\

BaRe-Mem
& 76.2 & 75.4 & 74.1 & 72.7 & 72.0
& \textbf{75.0} & \textbf{72.4} & \textbf{70.8} & \textbf{70.2} & \textbf{69.3} \\

\midrule

Consultation ratio
& 86\% & 85\% & 85\% & 86\% & 85\%
& 86\% & 78\% & 69\% & 53\% & 15\% \\

\bottomrule
\end{tabular}
}
\end{table}

\begin{table}[!h]
\centering
\caption{
Accuracy (\%) of consultation methods with Qwen3-8B as the central model under increasing misleading information ratios.
Results are reported for the capability-supported and capability-challenging regimes.
Bold numbers indicate the best performance in each column.
}
\label{tab:qwen3-8b-capability}
\resizebox{\textwidth}{!}{
\begin{tabular}{l|ccccc|ccccc}
\toprule
\multirow{2}{*}{\textbf{Method}}
& \multicolumn{5}{c|}{\textbf{Capability-supported}}
& \multicolumn{5}{c}{\textbf{Capability-challenging}} \\
\cmidrule(lr){2-6} \cmidrule(lr){7-11}
& \textbf{0\%} & \textbf{25\%} & \textbf{50\%} & \textbf{75\%} & \textbf{100\%}
& \textbf{0\%} & \textbf{25\%} & \textbf{50\%} & \textbf{75\%} & \textbf{100\%} \\
\midrule

No consultation
& 72.9 & 72.9 & 72.9 & 72.9 & 72.9
& 65.6 & 65.6 & 65.6 & 65.6 & 65.6 \\

Question + Peers
& 75.7 & 75.1 & 74.3 & 73.6 & 72.3
& 67.9 & 61.6 & 54.3 & 46.7 & 40.0 \\

Majority vote (advisors)
& 64.3 & 53.0 & 32.5 & 14.6 & 9.4
& 64.1 & 48.7 & 24.5 & 5.9 & 1.2 \\

Majority vote (advisors + own)
& 68.9 & 62.7 & 47.6 & 27.9 & 19.3
& 66.5 & 55.5 & 33.3 & 9.7 & 2.7 \\

Debate (2 rounds)
& 74.8 & 74.8 & 74.6 & 74.4 & 73.8
& 67.0 & 63.9 & 60.3 & 55.7 & 51.3 \\

Advisors + memory
& \textbf{78.1} & \textbf{77.0} & \textbf{76.3} & \textbf{74.8} & \textbf{74.0}
& 74.2 & 70.1 & 66.9 & 61.0 & 46.2 \\

BaRe-Mem
& 77.4 & 76.6 & 76.1 & 74.6 & 73.7
& \textbf{74.9} & \textbf{71.4} & \textbf{70.0} & \textbf{68.7} & \textbf{67.7} \\

\midrule

Consultation ratio
& 86\% & 85\% & 85\% & 84\% & 83\%
& 89\% & 80\% & 70\% & 52\% & 14\% \\

\bottomrule
\end{tabular}
}
\end{table}

\begin{figure}[!h]
\centering
\includegraphics[width=0.9\linewidth]{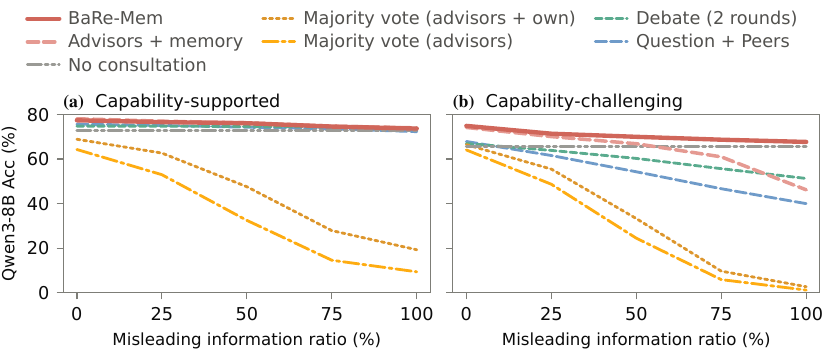}
\caption{
Accuracy of Qwen3-8B as the central model under increasing misleading information ratios.
\textbf{Left:} capability-supported regime; \textbf{right:} capability-challenging regime.
We compare BaRe-Mem with \textit{Advisors + memory}, \textit{No consultation}, \textit{Question + Peers}, \textit{Debate (2 rounds)}, and two majority voting baselines.
}
\vspace{-0.8em}
\label{fig:qwen3-8b-misleading}
\end{figure}

\begin{table}[!h]
\centering
\caption{
Accuracy (\%) of consultation methods with Qwen3-14B as the central model
under increasing misleading information ratios.
Results are reported for the capability-supported and capability-challenging regimes.
Bold numbers indicate the best performance in each column.
}
\label{tab:qwen3-14b-capability-comparison}
\resizebox{\textwidth}{!}{
\begin{tabular}{l|ccccc|ccccc}
\toprule
\multirow{2}{*}{\textbf{Method}}
& \multicolumn{5}{c|}{\textbf{Capability-supported}}
& \multicolumn{5}{c}{\textbf{Capability-challenging}} \\
\cmidrule(lr){2-6}
\cmidrule(lr){7-11}
& \textbf{0\%}
& \textbf{25\%}
& \textbf{50\%}
& \textbf{75\%}
& \textbf{100\%}
& \textbf{0\%}
& \textbf{25\%}
& \textbf{50\%}
& \textbf{75\%}
& \textbf{100\%} \\
\midrule

No consultation
& 72.9 & 72.9 & 72.9 & 72.9 & 72.9
& 66.5 & 66.5 & 66.5 & 66.5 & 66.5 \\

Question + Peers
& 76.6 & 75.9 & 75.3 & 73.9 & 73.2
& 69.8 & 65.6 & 60.4 & 55.2 & 50.6 \\

Majority vote (advisors)
& 64.3 & 53.0 & 32.5 & 14.6 & 9.4
& 64.1 & 48.7 & 24.5 & 5.9 & 1.2 \\

Majority vote (advisors + own)
& 68.5 & 62.6 & 47.5 & 27.8 & 19.4
& 66.4 & 55.8 & 33.6 & 9.9 & 2.8 \\

Debate (2 rounds)
& 74.9 & 74.7 & 73.9 & 74.0 & 73.4
& 69.2 & 66.7 & 62.6 & 58.9 & 55.4 \\

Advisors + memory
& \textbf{78.0}
& \textbf{77.4}
& \textbf{76.2}
& \textbf{75.2}
& \textbf{74.5}
& 75.6 & 72.1 & 69.5 & 65.2 & 54.7 \\

BaRe-Mem
& 77.7 & 77.3 & 76.0 & 75.1 & \textbf{74.5}
& \textbf{76.1}
& \textbf{72.8}
& \textbf{71.6}
& \textbf{69.9}
& \textbf{68.6} \\

\midrule

Consultation ratio
& 87\% & 86\% & 85\% & 85\% & 85\%
& 90\% & 83\% & 76\% & 61\% & 21\% \\

\bottomrule
\end{tabular}
}
\end{table}

\begin{figure}[!h]
\centering
\includegraphics[width=0.9\linewidth]{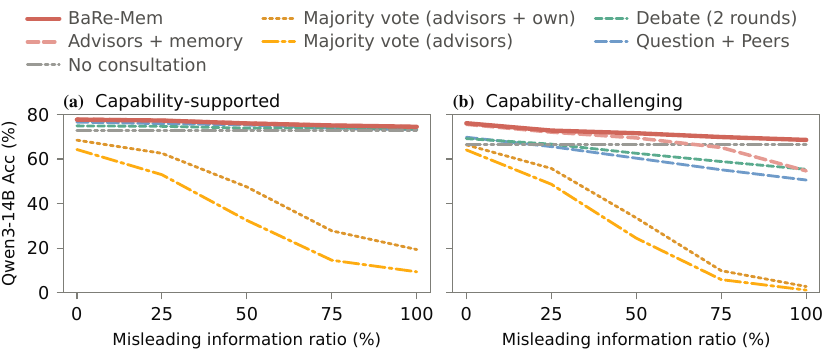}
\caption{
Accuracy of Qwen3-14B as the central model under increasing misleading information ratios.
\textbf{Left:} capability-supported regime; \textbf{right:} capability-challenging regime.
We compare BaRe-Mem with \textit{Advisors + memory}, \textit{No consultation}, \textit{Question + Peers}, \textit{Debate (2 rounds)}, and two majority voting baselines.
}
\vspace{-0.8em}
\label{fig:qwen3-14b-misleading}
\end{figure}

\begin{table}[!h]
\centering
\caption{
Accuracy (\%) of consultation methods with Qwen2.5-7B as the central model
under increasing misleading information ratios.
Results are reported for the capability-supported and capability-challenging regimes.
Bold numbers indicate the best performance in each column.
}
\label{tab:qwen2.5-7b-capability-comparison}
\resizebox{\textwidth}{!}{
\begin{tabular}{l|ccccc|ccccc}
\toprule
\multirow{2}{*}{\textbf{Method}}
& \multicolumn{5}{c|}{\textbf{Capability-supported}}
& \multicolumn{5}{c}{\textbf{Capability-challenging}} \\
\cmidrule(lr){2-6}
\cmidrule(lr){7-11}
& \textbf{0\%}
& \textbf{25\%}
& \textbf{50\%}
& \textbf{75\%}
& \textbf{100\%}
& \textbf{0\%}
& \textbf{25\%}
& \textbf{50\%}
& \textbf{75\%}
& \textbf{100\%} \\
\midrule

No consultation
& 70.6
& 70.6
& 70.6
& \textbf{70.6}
& \textbf{70.6}
& 59.1
& 59.1
& 59.1
& 59.1
& 59.1 \\

Question + Peers
& 71.3
& 70.3
& 69.2
& 67.8
& 67.1
& 65.9
& 59.3
& 52.9
& 47.3
& 43.3 \\

Majority vote (advisors)
& 64.3
& 53.0
& 32.5
& 14.6
& 9.4
& 64.1
& 48.7
& 24.5
& 5.9
& 1.2 \\

Majority vote (advisors + own)
& 68.4
& 62.3
& 47.1
& 27.3
& 18.9
& 64.7
& 53.5
& 31.5
& 8.8
& 2.2 \\

Debate (2 rounds)
& 70.5
& 70.3
& 69.9
& 69.2
& 68.2
& 63.1
& 58.6
& 53.7
& 49.1
& 45.5 \\

Advisors + memory
& \textbf{75.5}
& \textbf{74.0}
& \textbf{72.9}
& \textbf{70.6}
& 69.2
& 71.4
& 68.0
& 64.7
& 58.7
& 48.4 \\

BaRe-Mem
& 74.9
& 73.6
& 72.5
& 70.5
& 69.3
& \textbf{71.5}
& \textbf{68.3}
& \textbf{66.1}
& \textbf{63.5}
& \textbf{61.9} \\

\midrule

Consultation ratio
& 84\%
& 81\%
& 82\%
& 83\%
& 77\%
& 90\%
& 83\%
& 75\%
& 54\%
& 19\% \\

\bottomrule
\end{tabular}
}
\end{table}

\begin{figure}[!h]
\centering
\includegraphics[width=0.9\linewidth]{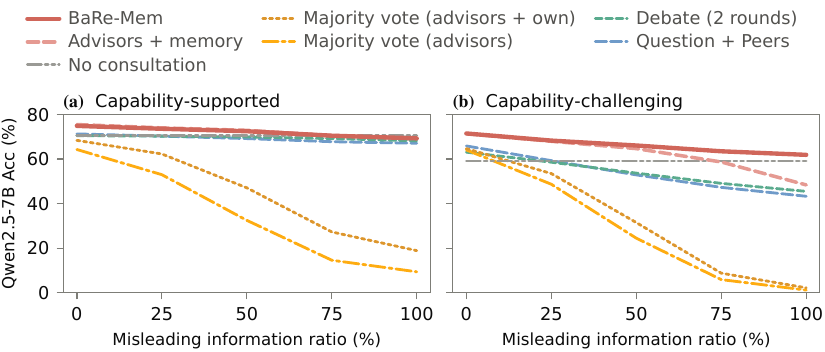}
\caption{
Accuracy of Qwen2.5-7B as the central model under increasing misleading information ratios.
\textbf{Left:} capability-supported regime; \textbf{right:} capability-challenging regime.
We compare BaRe-Mem with \textit{Advisors + memory}, \textit{No consultation}, \textit{Question + Peers}, \textit{Debate (2 rounds)}, and two majority voting baselines.
}
\vspace{-0.8em}
\label{fig:qwen2.5-7b-misleading}
\end{figure}

\begin{table}[!h]
\centering
\caption{
Accuracy (\%) of consultation methods with Ministral-8B as the central model
under increasing misleading information ratios.
Results are reported for the capability-supported and capability-challenging regimes.
Bold numbers indicate the best performance in each column.
}
\label{tab:ministral-8b-capability-comparison}
\resizebox{\textwidth}{!}{
\begin{tabular}{l|ccccc|ccccc}
\toprule
\multirow{2}{*}{\textbf{Method}}
& \multicolumn{5}{c|}{\textbf{Capability-supported}}
& \multicolumn{5}{c}{\textbf{Capability-challenging}} \\
\cmidrule(lr){2-6}
\cmidrule(lr){7-11}
& \textbf{0\%}
& \textbf{25\%}
& \textbf{50\%}
& \textbf{75\%}
& \textbf{100\%}
& \textbf{0\%}
& \textbf{25\%}
& \textbf{50\%}
& \textbf{75\%}
& \textbf{100\%} \\
\midrule

No consultation
& 56.5 & 56.5 & 56.5 & 56.5 & 56.5
& 58.6 & 58.6 & 58.6 & 58.6 & 58.6 \\

Question + Peers
& 64.2 & 63.1 & 61.1 & 58.5 & 56.9
& 64.0 & 54.8 & 41.6 & 27.4 & 18.0 \\

Majority vote (advisors)
& 64.3 & 53.0 & 32.5 & 14.6 & 9.4
& 64.1 & 48.7 & 24.5 & 5.9 & 1.2 \\

Majority vote (advisors + own)
& 67.1 & 59.7 & 43.6 & 24.9 & 17.1
& 64.2 & 53.3 & 31.5 & 8.9 & 2.2 \\

Debate (2 rounds)
& 61.8 & 60.8 & 59.0 & 57.0 & 56.7
& 61.5 & 56.5 & 49.7 & 41.3 & 34.4 \\

Advisors + memory
& \textbf{70.8}
& \textbf{67.8}
& \textbf{64.6}
& 60.3
& 60.0
& \textbf{70.8}
& 65.8
& 61.5
& 52.3
& 26.0 \\

BaRe-Mem
& 70.2
& 67.2
& 64.4
& \textbf{60.8}
& \textbf{61.0}
& \textbf{70.8}
& \textbf{66.5}
& \textbf{64.2}
& \textbf{62.7}
& \textbf{60.6} \\

\midrule

Consultation ratio
& 85\%
& 83\%
& 81\%
& 72\%
& 71\%
& 89\%
& 79\%
& 65\%
& 40\%
& 9\% \\

\bottomrule
\end{tabular}
}
\end{table}

\begin{figure}[!h]

\centering
\includegraphics[width=0.9\linewidth]{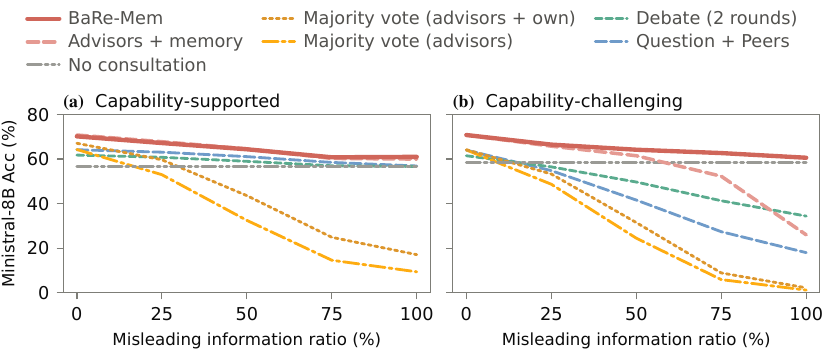}
\caption{Accuracy under increasing misleading-advice ratios with Ministral-8B as the central model. The left and right panels show the capability-supported and capability-challenging regimes, respectively. We compare BaRe-Mem, \textit{Advisors + memory}, \textit{No consultation}, \textit{Question + Peers}, \textit{Debate (2 rounds)}, and two majority-vote variants.}
\vspace{-0.8em}
\label{fig:exps1}
\end{figure}

\clearpage
\begin{table}[H]
\centering
\caption{
Accuracy (\%) of consultation methods with Phi-4 as the central model
under increasing misleading information ratios.
Results are reported for the capability-supported and capability-challenging regimes.
Bold numbers indicate the best performance in each column.
}
\label{tab:phi-4-capability-comparison}
\resizebox{\textwidth}{!}{
\begin{tabular}{l|ccccc|ccccc}
\toprule
\multirow{2}{*}{\textbf{Method}}
& \multicolumn{5}{c|}{\textbf{Capability-supported}}
& \multicolumn{5}{c}{\textbf{Capability-challenging}} \\
\cmidrule(lr){2-6}
\cmidrule(lr){7-11}
& \textbf{0\%}
& \textbf{25\%}
& \textbf{50\%}
& \textbf{75\%}
& \textbf{100\%}
& \textbf{0\%}
& \textbf{25\%}
& \textbf{50\%}
& \textbf{75\%}
& \textbf{100\%} \\
\midrule

No consultation
& 70.2
& 70.2
& 70.2
& 70.2
& 70.2
& 63.4
& 63.4
& 63.4
& 63.4
& 63.4 \\

Question + Peers
& 71.5
& 72.1
& 71.6
& 70.8
& 69.9
& 65.8
& 62.4
& 57.9
& 52.5
& 48.3 \\

Majority vote (advisors)
& 64.3
& 53.0
& 32.5
& 14.6
& 9.4
& 64.1
& 48.7
& 24.5
& 5.9
& 1.2 \\

Majority vote (advisors + own)
& 67.9
& 61.5
& 46.3
& 26.7
& 18.3
& 65.1
& 54.3
& 32.1
& 9.0
& 2.3 \\

Debate (2 rounds)
& 68.0
& 68.8
& 69.4
& 69.6
& 68.6
& 64.8
& 62.6
& 59.3
& 55.2
& 51.4 \\

Advisors + memory
& 76.3
& 75.0
& 73.6
& 72.9
& 71.6
& \textbf{71.3}
& 68.4
& 65.6
& 61.8
& 52.0 \\

BaRe-Mem
& \textbf{76.5}
& \textbf{75.2}
& \textbf{73.9}
& \textbf{73.0}
& \textbf{71.8}
& 71.2
& \textbf{68.5}
& \textbf{66.8}
& \textbf{65.3}
& \textbf{65.0} \\

\midrule

Consultation ratio
& 85\%
& 84\%
& 83\%
& 84\%
& 83\%
& 85\%
& 79\%
& 71\%
& 57\%
& 18\% \\

\bottomrule
\end{tabular}
}
\end{table}
       
\FloatBarrier
\section{Agent Team}
\subsection{Agent Team Pipeline.}
\begin{figure}[!h]
\centering
\includegraphics[width=1.0\linewidth]{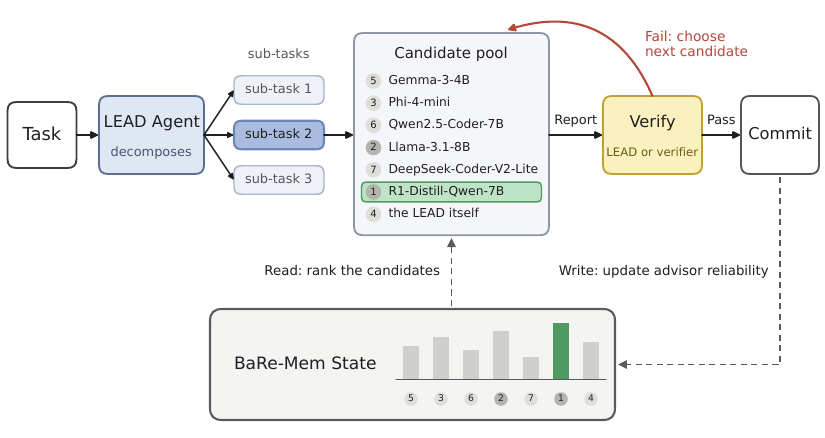}
\caption{
Agent team pipeline with BaRe-Mem. The lead agent decomposes a task into sub-tasks and queries the reliability memory to rank candidate workers. Each returned report is verified by the lead agent or an external verifier: accepted reports are committed, while rejected reports trigger the next candidate. Verification outcomes are then written back to update advisor reliability.
}
\vspace{-0.8em}
\label{fig:agent-team-pipeline}
\end{figure}

Figure~\textcolor{blue}{\textbf{\ref{fig:agent-team-pipeline}}}
illustrates the agent team workflow.
The lead agent first decomposes each task into sub-tasks and uses
BaRe-Mem to rank the candidate workers for each sub-task.
The highest rank worker is allocated first. If its report is rejected,
the lead agent retries with the next worker in the same ranking.
In this setting, BaRe-Mem is used for worker routing rather
than reweighting a fixed set of responses, deciding both which worker is selected first and which worker is tried next. To isolate the effect of worker routing from task decomposition quality,
we use the MuSiQue~\citep{trivedi2022musique} sub-tasks provided by the dataset, so all experimental
differences arise from how these sub-tasks are assigned to workers.

\textbf{Checking and Feedback Protocol.}
We distinguish the check used to control retries from the feedback used
to update reliability memory. We consider three checking settings:
\textbf{\textit{No check}}, where the first report is directly committed;
\textbf{\textit{Lead agent check}}, where the lead agent decides whether to
accept the report or retry; and \textbf{\textit{Dataset verifier check}}, where
the dataset evaluator makes this decision.
In all three settings, after a sub-task is committed, the gold
correctness labels of all workers actually queried for that sub-task
are written to memory, while unqueried workers receive no update.
Thus, checking determines the execution path of the current sub-task,
whereas verified feedback updates worker reliability for subsequent
routing decisions. Memory is updated after each sub-task, allowing later
sub-tasks to benefit from earlier verified outcomes.

\subsection{Task Solved Accuracy}
\begin{figure}[!h]
\centering
\includegraphics[width=1.0\linewidth]{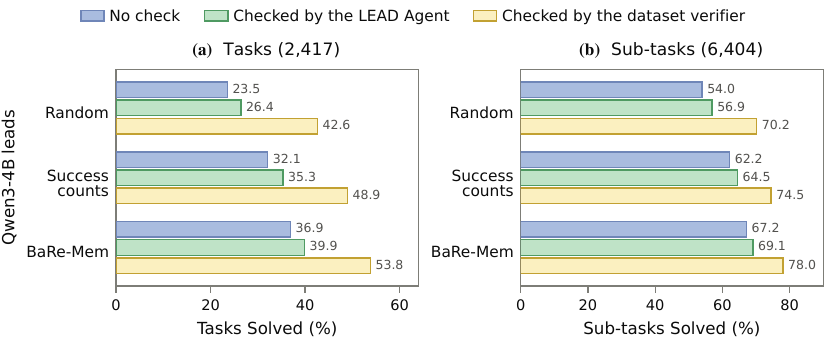}
\caption{
Agent team performance on MuSiQue with Qwen3-4B as the lead agent.
\textbf{(a)} Task completion; \textbf{(b)} sub-task completion.
We compare random routing, routing by historical success counts, and BaRe-Mem under three report checking settings: no check, lead agent check, and dataset verifier check.
All settings provide gold feedback for queried reports after each sub-task is committed.
}
\vspace{-0.8em}
\label{fig:agent-team-qwen3-4b}
\end{figure}

\begin{figure}[!h]
\centering
\includegraphics[width=1.0\linewidth]{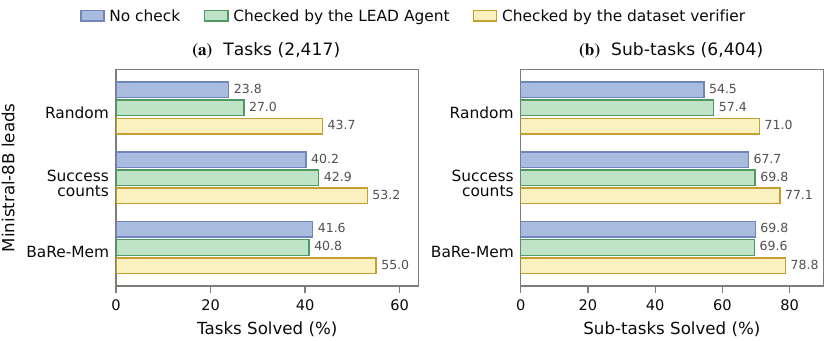}
\caption{
Agent team performance on MuSiQue with Ministral-8B as the lead agent.
\textbf{(a)} Task completion; \textbf{(b)} sub-task completion.
We compare random routing, routing by historical success counts, and BaRe-Mem under three report checking settings: no check, lead agent check, and dataset verifier check.
All settings provide gold feedback for queried reports after each sub-task is committed.
}
\vspace{-0.8em}
\label{fig:agent-team-ministral-8b}
\end{figure}

Figures~\textcolor{blue}{\textbf{\ref{fig:agent-team-qwen3-4b}}}--\textcolor{blue}{\textbf{\ref{fig:agent-team-ministral-8b}}}
report task and sub-task completion for six lead models:
\qwen Qwen3-4B, \qwen Qwen3-8B, \qwen Qwen3-14B, \qwen Qwen2.5-7B, \ministra Ministral-8B,
and \microsoft Phi-4. These results examine how reliability memory
supports worker assignment and subsequent retries
across different central models.

\textbf{Obs.1. Contextual reliability improves initial worker selection.}
The \textbf{\textit{No check}} setting commits the first worker's report
without retrying, directly evaluating the initial assignment.
In this setting, \textbf{BaRe-Mem achieves higher sub-task
completion than routing randomly or by historical success counts for all
six lead models}. The improvements are $5.0\%$, $1.4\%$, and $3.8\%$ points for \qwen Qwen3-4B, Qwen3-8B, and Qwen3-14B,
respectively, and $5.6\%$, $2.1\%$, and $6.9\%$ points for
\qwen Qwen2.5-7B, \ministra Ministral-8B, and \microsoft Phi-4.
Relative to random routing, it also improves task completion
by $13.4\%$--$17.8\%$ points across the six models.
Because these gains arise without retries, they support the
use of reliability conditioned on the current sub-task to
identify a suitable worker at the initial assignment.
The comparison with historical success counts further
indicates that the benefit extends beyond simply retaining
aggregate records of past success.

\textbf{Obs.2. Reliability ranking complements verification in Agent Team}
Under \textit{\textbf{Dataset verifier check}}, \textbf{BaRe-Mem
achieves the highest task and sub-task completion for every
lead model}. Task completion reaches $51.9\%$--$55.0\%$,
exceeding historical success counts by $1.8\%$--$4.9\%$
points and random routing by $9.1$--$11.4$ points.
Sub-task completion also improves over historical success
counts by $1.7\%$--$3.5\%$ points.
These comparisons use the same checking rule and worker
budget for each routing strategy, showing that access to a
verifier does not remove the benefit of our BaRe-Mem reliability worker
ranking. When the first report is rejected, the reliability
ranking also determines which worker is tried next.
The results support using the memory to guide
candidate ordering throughout the assignment and retry process.

\textbf{Obs.3. The effect of report checking depends on the lead model.}
Lead agent checking improves BaRe-Mem's task
completion over \textbf{\textit{No check}} for five of the six lead
models, with gains of $1.5\%$--$3.0\%$ points.
Its effect is not uniform: with Ministral-8B, task completion
changes from $41.6\%$ without checking to $40.8\%$ with
lead agent checking, whereas dataset verifier checking
raises it to $55.0\%$.
Across all six models, replacing lead agent checking with
dataset verifier checking improves BaRe-Mem's
task completion by $11.6\%$--$14.2\%$ points.
Together, these results distinguish the complementary roles
of routing and checking: reliability memory orders candidate
workers, while report checking determines when to proceed
to the next candidate.

\begin{figure}[!h]
\centering
\includegraphics[width=1.0\linewidth]{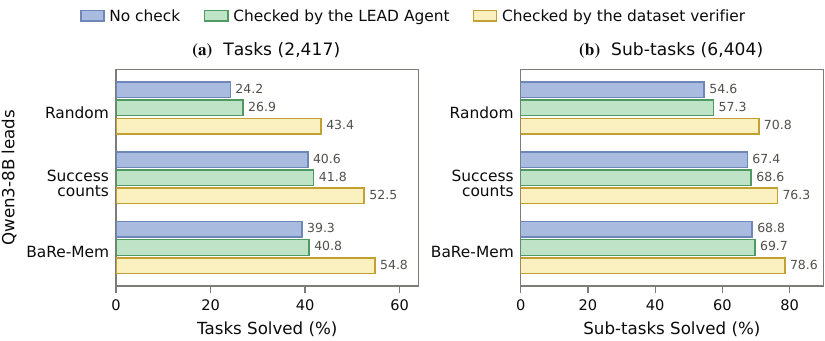}
\caption{
Agent team performance on MuSiQue with Qwen3-8B as the lead agent.
\textbf{(a)} Task completion; \textbf{(b)} sub-task completion.
We compare random routing, routing by historical success counts, and BaRe-Mem under three report checking settings: no check, lead agent check, and dataset verifier check.
All settings provide gold feedback for queried reports after each sub-task is committed.
}
\vspace{-0.8em}
\label{fig:agent-team-qwen3-8b}
\end{figure}

\begin{figure}[!h]
\centering
\includegraphics[width=1.0\linewidth]{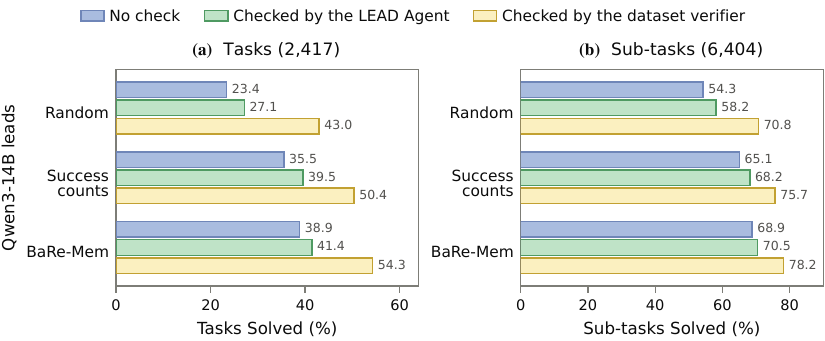}
\caption{
Agent team performance on MuSiQue with Qwen3-14B as the lead agent.
\textbf{(a)} Task completion; \textbf{(b)} sub-task completion.
We compare random routing, routing by historical success counts, and BaRe-Mem under three report checking settings: no check, lead agent check, and dataset verifier check.
All settings provide gold feedback for queried reports after each sub-task is committed.
}
\vspace{-0.8em}
\label{fig:agent-team-qwen3-14b}
\end{figure}

\begin{figure}[!h]
\centering
\includegraphics[width=1.0\linewidth]{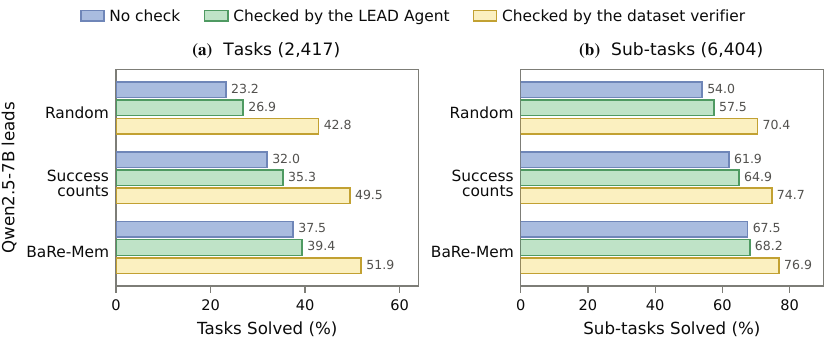}
\caption{
Agent team performance on MuSiQue with Qwen2.5-7B as the lead agent.
\textbf{(a)} Task completion; \textbf{(b)} sub-task completion.
We compare random routing, routing by historical success counts, and BaRe-Mem under three report checking settings: no check, lead agent check, and dataset verifier check.
All settings provide gold feedback for queried reports after each sub-task is committed.
}
\vspace{-0.8em}
\label{fig:agent-team-qwen2.5-7b}
\end{figure}

\begin{figure}[H]
\centering
\includegraphics[width=1.0\linewidth]{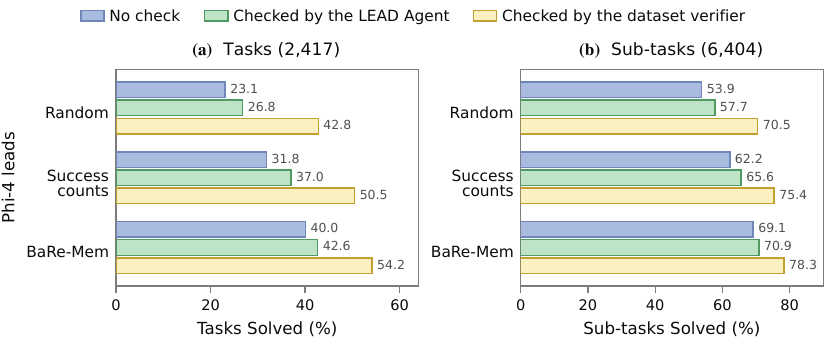}
\caption{
Agent team performance on MuSiQue with Phi-4 as the lead agent.
\textbf{(a)} Task completion; \textbf{(b)} sub-task completion.
We compare random routing, routing by historical success counts, and BaRe-Mem under three report checking settings: no check, lead agent check, and dataset verifier check.
All settings provide gold feedback for queried reports after each sub-task is committed.
}
\vspace{-0.8em}
\label{fig:agent-team-phi4}
\end{figure}

\FloatBarrier
\subsection{Agent Team Sweep}

\begin{figure}[!h]
\centering
\includegraphics[width=1.0\linewidth]{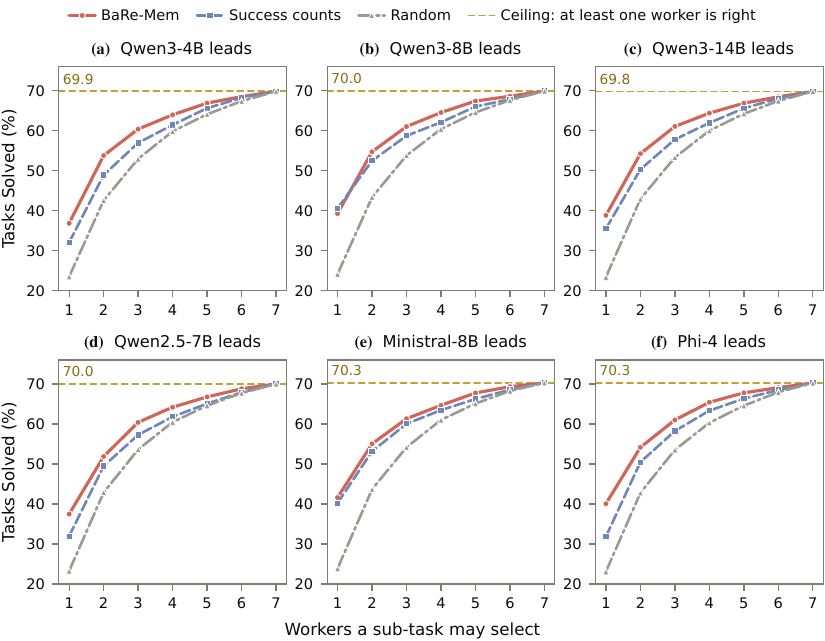}
\caption{
Task completion on MuSiQue as the maximum number of workers tried per sub-task increases.
We compare BaRe-Mem, routing by historical success counts, and random routing.
The dashed line marks the ceiling obtained when at least one worker can solve the sub-task.
Panels (a)--(f) use Qwen3-4B, Qwen3-8B, Qwen3-14B, Qwen2.5-7B, Ministral-8B, and Phi-4 as the lead agent, respectively.
}
\vspace{-0.8em}
\label{fig:agent-team-sweep-task}
\end{figure}

We examine how quickly each routing strategy can identify a worker capable of solving the current sub-task for all the central models following the setting in the main text Sec. \textcolor{blue}{\textbf{~\ref{real-world_application}}}. As the worker budget increases, all methods gradually approach the same ceiling once all seven workers have been tried. This is expected, because the candidate pool is identical across routing strategies. The remaining difference is therefore not \textit{whether} a solvable worker exists, but \textit{how early} that worker is ranked and selected.

\begin{figure}[!h]
\centering
\includegraphics[width=1.0\linewidth]{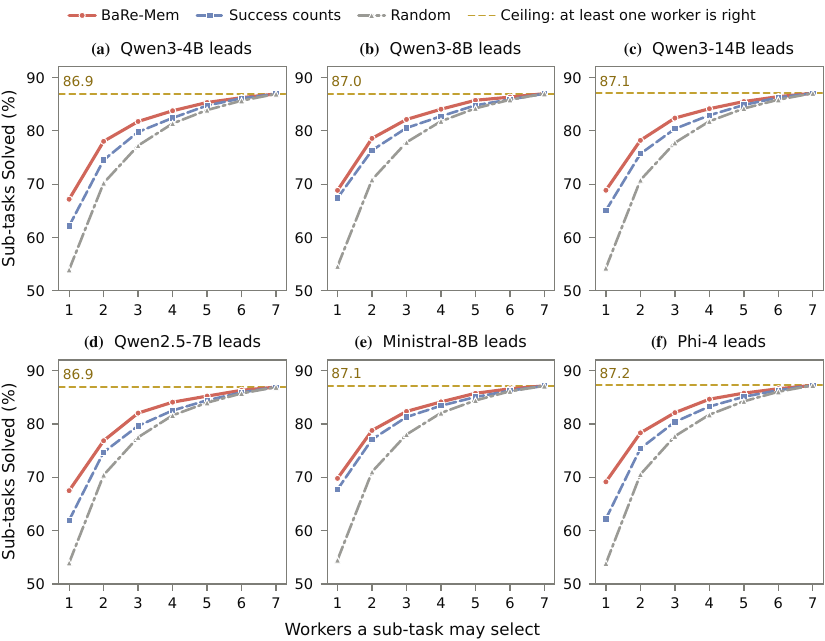}
\caption{
Sub-task completion on MuSiQue as the maximum number of workers tried per sub-task increases.
We compare BaRe-Mem, routing by historical success counts, and random routing.
The dashed line marks the ceiling given by the fraction of sub-tasks for which at least one worker is correct.
Panels (a)--(f) use Qwen3-4B, Qwen3-8B, Qwen3-14B, Qwen2.5-7B, Ministral-8B, and Phi-4 as the lead agent, respectively.
}
\vspace{-0.8em}
\label{fig:agent-team-sweep-subtask}
\end{figure}

\textbf{Obs. 1. BaRe-Mem identifies capable workers earlier.}
Figures~\textcolor{blue}{\textbf{\ref{fig:agent-team-sweep-task}}}
and~\textcolor{blue}{\textbf{\ref{fig:agent-team-sweep-subtask}}}
show that \textbf{BaRe-Mem consistently achieves higher task and
sub-task completion with fewer worker calls across all six lead models}.
The advantage is largest when only a small number of workers may be
tried. With a single worker per sub-task, BaRe-Mem already
outperforms routing by historical success counts for every lead model.
As the worker budget increases, all methods gradually approach the same
ceiling because they eventually access the same candidate pool.
Therefore, the gap at smaller budgets reflects how early a capable
worker is ranked and selected. BaRe-Mem reaches a larger
fraction of the achievable performance with the same worker budget.

\textbf{Obs. 2. Contextual reliability improves worker ranking.}
The comparison with historical success counts isolates the value of
contextual reliability. Success counts assign each worker a reliability
based only on its aggregate past performance, whereas BaRe-Mem
conditions worker reliability on the current sub-task. \textbf{The consistent
advantage of BaRe-Mem across all six lead models} therefore
shows that worker competence cannot be adequately represented by a
single source-level success rate. Conditioning reliability on the
current sub-task yields a more informative worker ordering, supporting
the same instance-specific reliability principle used in advisor
consultation.

\clearpage
\subsection{Adapt Speed}
We finally examine how quickly BaRe-Mem learns useful worker
reliability from the online feedback stream. Following the original
sub-task order, we partition the 6,404 MuSiQue sub-tasks into eight
consecutive blocks of approximately 800 sub-tasks each and measure the
accuracy of the \textbf{initial worker} selected in each block. This
directly measures how the quality of the learned worker ranking evolves
as verified interactions accumulate along the stream.

\begin{figure}[!h]
\centering
\includegraphics[width=1.0\linewidth]{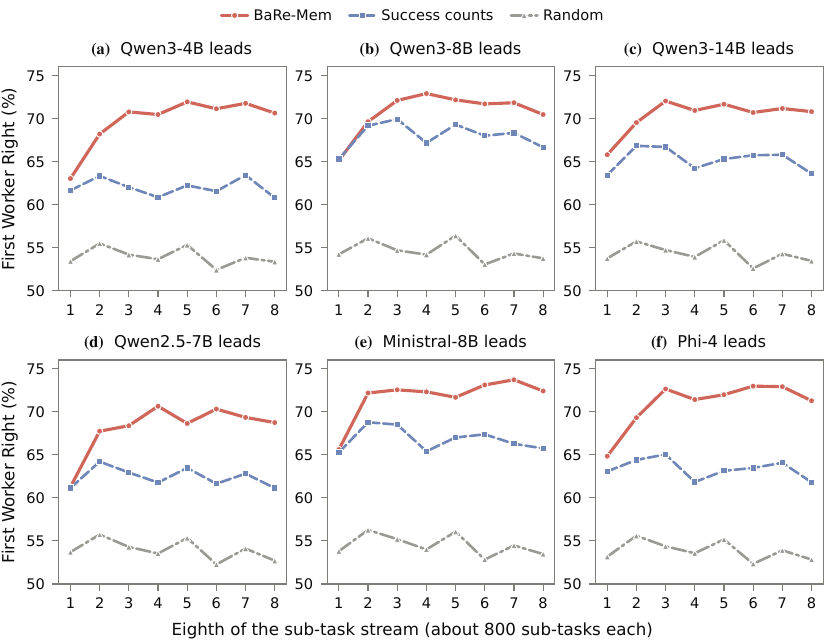}
\caption{
Initial worker accuracy over the MuSiQue sub-task stream.
The x-axis denotes position in the sub-task stream, grouped into eight consecutive blocks of approximately 800 sub-tasks each.
We compare BaRe-Mem, routing by historical success counts,
and random routing.
Panels (a)--(f) use Qwen3-4B, Qwen3-8B, Qwen3-14B, Qwen2.5-7B,
Ministral-8B, and Phi-4 as the lead agent, respectively.
}
\vspace{-0.8em}
\label{fig:agent-team-adaptation}
\end{figure}

\textbf{BaRe-Mem adapts rapidly from early feedback.}
Figure~\textcolor{blue}{\textbf{\ref{fig:agent-team-adaptation}}}
shows that initial worker accuracy rises quickly during the early part of
the stream for all six lead models and then remains relatively stable.
Historical success counts also improve as feedback accumulates, but
\textbf{BaRe-Mem separates from this baseline early and maintains higher
initial worker accuracy throughout the stream}. In contrast, random routing
remains nearly unchanged. These results show that \textbf{BaRe-Mem} can
rapidly extract useful reliability information from early interactions.

This adaptation pattern is also consistent with the Bayesian update in
Section~\textcolor{blue}{\textbf{\ref{app:kalman_gain}}}: when posterior
uncertainty is high, the Kalman gain gives early verified outcomes greater
influence, while accumulated evidence reduces uncertainty and stabilizes
subsequent updates. The resulting behavior allows \textbf{BaRe-Mem} to
adapt quickly without requiring a separate retraining stage.
\end{document}